# The Latin Substrate: How Language Models Represent and Mediate Script Choice

**Daniil Gurgurov**[D, S, *] **Alan Saji**[N, I, *] **Katharina Trinley**[S, H*]
**Josef van Genabith**[D, S] **Simon Ostermann**[D, S]
[D] German Research Center for AI (DFKI) [S] Saarland University
[H] University of Hamburg [I] Indian Institute of Technology, Madras [N] Nilekani Centre at AI4Bharat
daniil.gurgurov@dfki.de, alansaji2001@gmail.com, katharina.trinley@uni-hamburg.de

## Abstract

Many languages are written in multiple scripts, requiring large language models (LLMs) to generate equivalent linguistic content in distinct orthographic forms. While prior work suggests that LLMs route information through shared latent representations, how they internally mediate script variation remains poorly understood.

We study this question by first examining per-layer output distributions with the logit lens, which reveals consistent latent romanization during transliteration, and then through representational and mechanistic analyses of script generation. At the *representational* level, we show that scripts of the same language become increasingly separable across layers and that a simple linear steering direction can flip a model's output script while largely preserving semantic content. The vector generalizes to writing systems unseen during construction: it flips non-Latin output to Latin, but maps Latin output into varied non-Latin scripts. At the *mechanistic* level, we localize a small set of late-layer attention heads that causally mediate script choice. These heads transfer across unrelated languages and writing systems, suggesting that script routing is implemented by language-agnostic components. Across both analyses, we observe a consistent directional asymmetry: non-Latin output is produced by a compact, identifiable gate, while Latin-script output emerges from diffuse contributions across the model. Overall, our findings hint that LLMs organize script variation around shared latent representations while exhibiting a privileged substrate toward Latin script.

## 1 Introduction

Many languages are written in multiple scripts (Bunčić, 2016; Vaid, 2022): Hindi appears in both Devanagari and Latin, Serbian in Cyrillic and Latin, and Urdu increasingly in both Perso-Arabic and Latin forms. Multilingual large language models (LLMs) trained on such mixed-script data must therefore generate equivalent linguistic content across distinct orthographic systems (Zhao et al., 2024; Yang et al., 2025). This includes transliteration, the task of converting text between scripts while preserving the underlying language (Jayakumar et al., 2026). Romanization, the special case of transliterating into Latin script, is the most widely used variant (Ma et al., 2025).

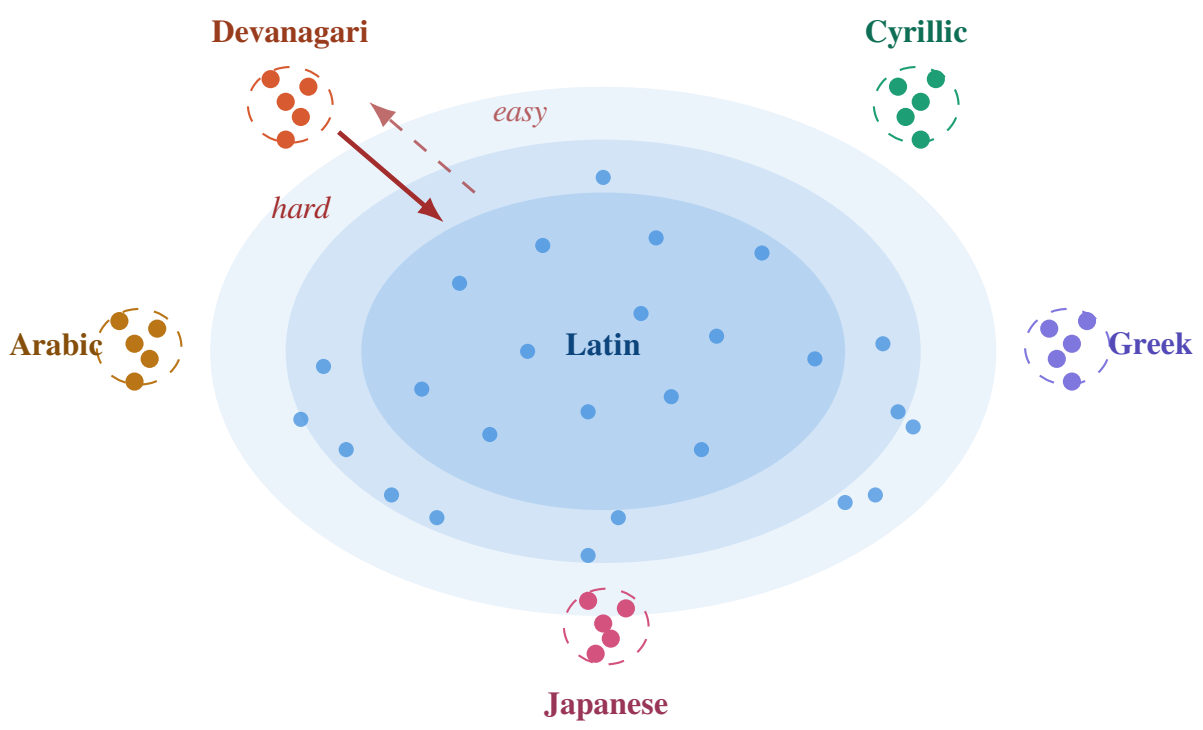

Figure 1: Schematic of script representations in LLMs. Latin script occupies a broad region of the representation space, while non-Latin scripts form more localized clusters. The schematic illustrates the directional asymmetry observed in both our vector-steering and head-patching interventions: steering toward non-Latin scripts targets a localized region, while steering toward Latin targets a more diffuse one.

How LLMs internally represent and mediate multiple scripts of the same language remains poorly understood. Prior work suggests that representations are at least partially shared across scripts. Romanizing low-resource languages improves downstream performance in multilingual models (Purkayastha et al., 2023; J et al., 2024). During translation involving non-Latin languages, romanized tokens emerge in intermediate layers, a phenomenon referred to as *latent romanization* (Saji et al., 2025). Sparse autoencoder analyses further show that the same sentence written in different scripts activates substantially overlapping features (Karne, 2026). Together, these observations suggest that script variation may be mediated through common internal representations rather than completely separate processing pathways.

* Equal contribution.

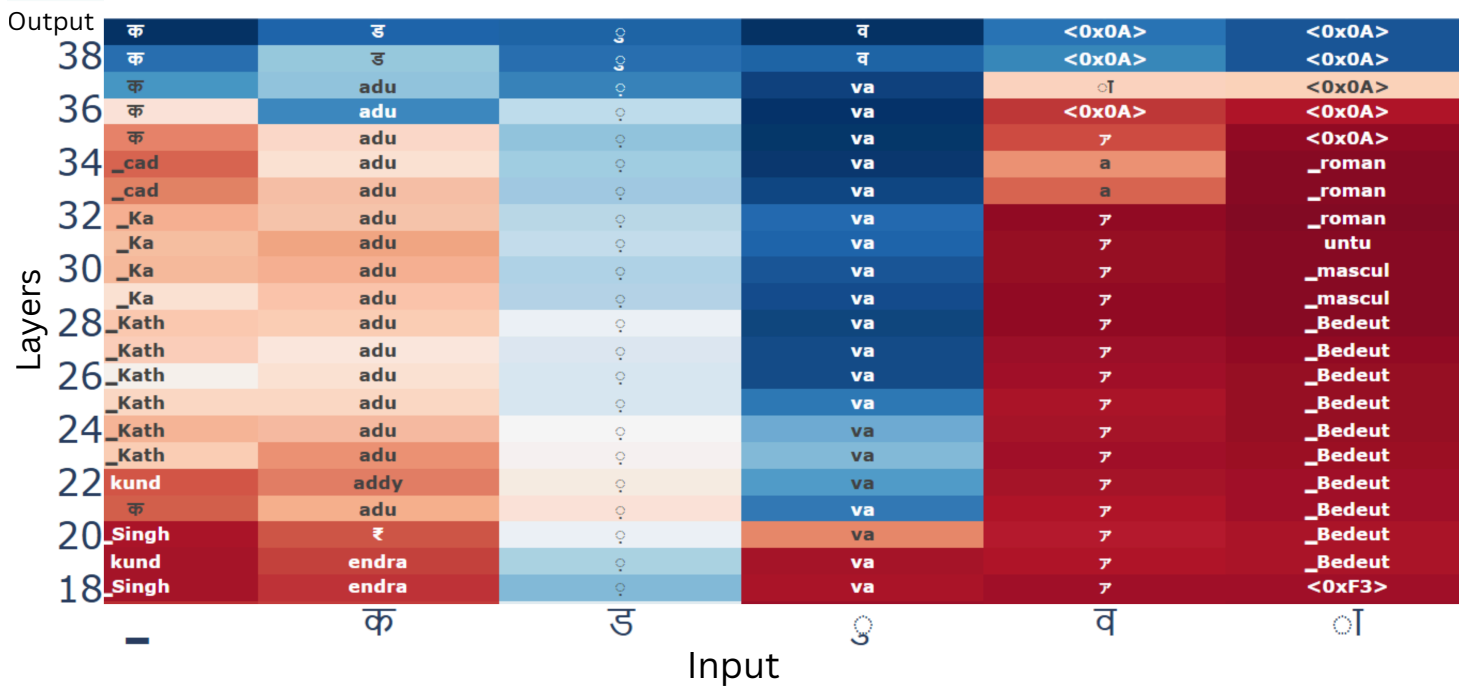

Figure 2: Logit lens visualization of `Llama-2-13B`[1] transliterating "tiger" from Malayalam to Devanagari. The plot shows the next-token distribution at each position (x-axis) across layers (y-axis), with the final output (कडुवा, "kaduva" in romanized form) taking shape from the bottom up. In the middle-to-upper layers (20–40), romanized subwords of the target word (क – ka; डु – adu; वा – va) appear before being transformed into the target script.

One way to study these representations is to examine the intermediate predictions of a model using the logit lens (Nostalgebraist, 2020). Figure 2 shows an example from `Llama-2-13B` (Touvron et al., 2023) as it transliterates the Malayalam word for "tiger" into Devanagari. In middle-to-upper layers, romanized subwords appear before the model produces the final Devanagari output. This suggests that the model may pass through a representation aligned with the romanized form before producing the target script; we quantify this across multiple scripts and models in Section 4. This observation extends prior work on latent romanization (Saji et al., 2025) to script transformation and motivates the central hypothesis of this paper.

**Central Hypothesis**

**Latin script acts as a privileged substrate for surface-language representations in LLMs.** LLMs appear to mediate script transformations through latent representations aligned with Latin script, while non-Latin outputs are produced through dedicated routing mechanisms.

We test this at three levels: a *decoding* analysis of what the model would output at intermediate layers, a *representational* analysis of whether scripts occupy separable regions of activation space and can be moved between them, and a *mechanistic* analysis of which internal components cause the choice of script. The paper makes the following four contributions:

- We quantitatively demonstrate that LLMs exhibit latent romanization during transliteration using logit lens analysis.
- We show that scripts of the same language become progressively separable across layers. A simple linear steering direction can reliably flip a model's output script while largely preserving meaning. This direction also transfers to writing systems that were not used to construct the steering vector.
- We localize a small set of late-layer attention heads that causally mediate script generation. Patching as few as five heads redirects the output script on naturalistic prompts, and the same heads transfer across unrelated languages and writing systems.
- Across the representational and mechanistic analyses, we observe a consistent directional asymmetry: non-Latin generation depends on compact, identifiable mechanisms, whereas Latin-script generation is more diffusely distributed across the model (Figure 1).

[1] We use `Llama-2-13B` for this qualitative example because its tokenizer represents Devanagari characters as single tokens, making layer-wise script transitions visually legible. Newer tokenizers (e.g., `Llama-3`) split Devanagari into byte-level pieces, producing fragmented logit lens output that obscures the visualization.

## 2 Related Work

**Transliteration as a cross-lingual bridge.** Transliteration creates anchor points that link meaning across languages whose different writing systems would otherwise prevent lexical overlap (Pires et al., 2019; Jayakumar et al., 2026). Romanization, in particular, has been shown to improve cross-lingual transfer in low-resource settings (Zoph et al., 2016; Johnson et al., 2017; Khemchandani et al., 2021; Purkayastha et al., 2023; J et al., 2024). These findings suggest that multilingual models rely on partially shared representations across languages and scripts.

**Latent representations across languages.** Prior work shows that intermediate representations are not necessarily tied to the surface form of the input. Wendler et al. (2024) find that LLMs process non-English inputs through latent English representations in intermediate layers. Closer to our setting, Saji et al. (2025) identify latent romanization, where intermediate layers transitorily represent information in romanized form before producing the target script. Zhong et al. (2025) further show that the privileged latent representation need not always be English: Japanese models alternate between English and Japanese depending on the target language.

**Localization of multilingual structure.** Recent mechanistic work has located multilingual structure inside model components. Tang et al. (2024) and Gurgurov et al. (2025) identify language-sensitive neurons concentrated in later layers, while Gurgurov et al. (2026) find language-specific steering vectors. Closest to our work, Shim et al. (2026) show that script identity is linearly encoded in speech-to-text models. However, similar work on *text* LLMs, particularly on how scripts are represented and routed during transliteration, remains scarce. Our work addresses this gap through a unified analysis spanning per-layer latent romanization, representation steering, and causal localization of script-mediating attention heads.[2]

## 3 Experimental Setup

**Models.** We analyze four open-source instruction-tuned LLMs spanning two model families and two scales: `Llama-3.1-8B`, `Llama-3.1-70B` (Grattafiori et al., 2024), `Aya-Expanse-8B`, and `Aya-Expanse-32B` (Dang et al., 2024). For the layer-wise output analysis, we also include `Llama-2-13B` (Touvron et al., 2023). Figures and qualitative examples in the main text use `Llama-3.1-8B`; similar results for the remaining models are reported in Appendices A, B, and C.

**Languages.** Our primary focus (Sections 5.1 and 6.1) is on Hindi (Devanagari/Latin) and Arabic (Arabic/Latin), using their naturally occurring romanized variants (Romanized Hindi and Arabizi, respectively). These languages span unrelated non-Latin writing systems while both supporting widely used romanized forms.

For our per-layer output distribution analysis (Section 4), we broaden the script set and transliterate Malayalam words into Devanagari, Arabic, Telugu, Cyrillic, and Greek; the appendix A also reports Greek as a source language. To evaluate cross-lingual generalization, we extend our representational analysis (Section 5.3) to 9 non-Latin and 9 Latin-script evaluation languages, and our mechanistic analysis (Section 6.3) to five cross-script languages (Russian, Marathi, Urdu, Greek, and Japanese).[3]

**Data.** For the per-layer output analysis, we use the simple word-level dataset from Saji et al. (2025), transliterating each word to the target scripts using `Gemini-3.1-Pro` (Google, 2026). For probing, steering vector construction, and head localization, we use 900 parallel native/romanized sentence pairs from FLORES-200 (Team et al., 2022). For steering and intervention evaluation, we use 70 held-out native/romanized prompt pairs from CLaS-Bench (Gurgurov et al., 2026). For cross-lingual attention head generalization experiments, we use a set of 10 open-ended questions generated with `Gemini-3.1-Pro` Google (2026) in 5 target languages, both in native and romanized scripts. Using separate datasets ensures that we do not evaluate on examples used to construct steering vectors or identify attention heads.

[2] Our code is available at https://github.com/d-gurgurov/Script-Mediation.

[3] We use a smaller set here as the head-patching requires parallel native/romanized prompts for each target language, and we restrict to languages with established romanization conventions for which such pairs can be reliably constructed.

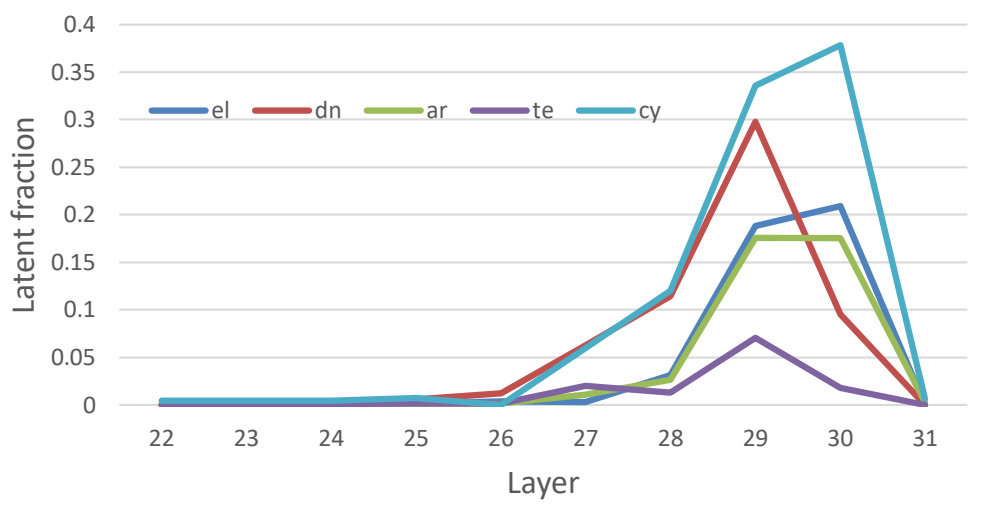

Figure 3: Distribution of romanized tokens across the last 10 layers of `Llama-3.1-8B` during transliteration with Malayalam as source, averaged over 130 samples. The target scripts are: Arabic (ar), Telugu (te), Devanagari (dn), Cyrillic (cy) and Greek (el).

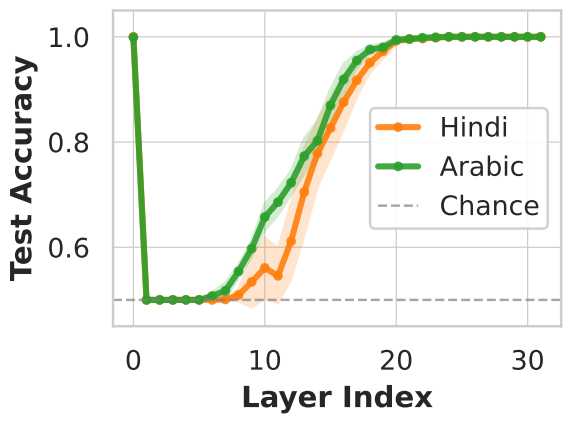

(a) Probe accuracy

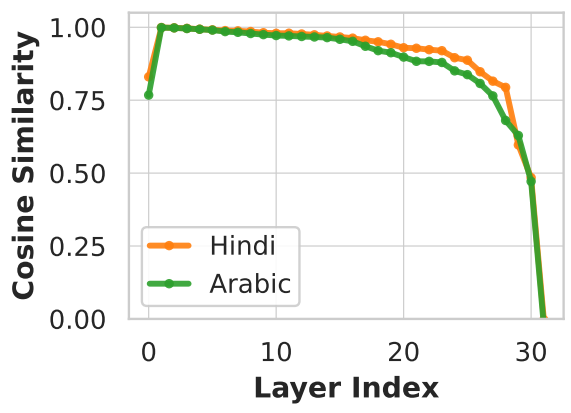

(b) Cos. similarity

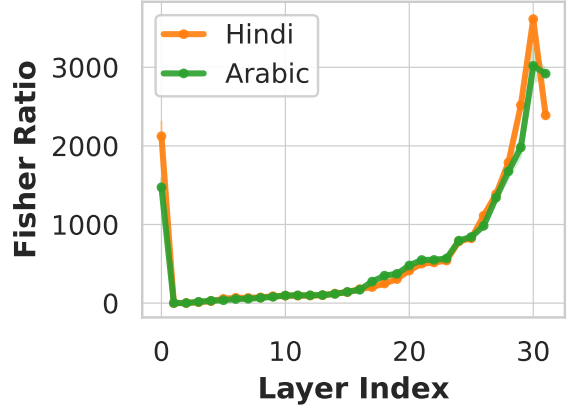

(c) Fisher ratio

Figure 4: Layer-wise probe accuracy, within-language similarity, and Fisher ratio for the two scripts of Hindi and Arabic. All metrics indicate that script separability emerges primarily in the later layers of `Llama-3.1-8B`.

## 4 Per-layer Output Analysis

The qualitative example in Figure 2 showed romanized subwords emerging in intermediate layers before the model committed to the target script. In this section, we quantify this observation for the transliteration task.

We use logit lens (Nostalgebraist, 2020), which applies the language modeling head to intermediate layers and decodes the resulting next-token distribution (see Appendix A.1). Following Saji et al. (2025), we quantify the *latent fraction*: the fraction of generation timesteps where romanized subwords appear in the top-1 prediction with probability $> 0.1$, averaged over samples (Appendix A.3). We prompt the model to transliterate a word given five in-context examples; an example prompt is depicted in Appendix A.2. We analyze the final ten layers of the LLMs, where coherent romanized representations are observed in the qualitative analysis.

Figure 3 shows the layer-wise latent fraction for `Llama-3.1-8B` on transliterations from Malayalam to five target scripts. Romanized representations appear at non-trivial rates in upper layers across all five scripts, and the same pattern holds across the remaining models and transliteration directions (Appendix A.4). The latent fraction is consistently higher for transliteration than for the translation, repetition, and cloze tasks reported by Saji et al. (2025).

This finding is consistent with the hypothesis that Latin acts as a shared substrate for script transformations. It also suggests that latent romanization in LLMs parallels how humans use pronunciation as a script-agnostic bridge for transliteration (Wellisch et al., 1978; DeFrancis, 1989; Coltheart et al., 2001; Rastle and Brysbaert, 2006; Rao et al., 2013).

## 5 Representation Analysis

We next ask whether script information is accessible at the representational level. If different scripts of the same language occupy separate subspaces, they should be linearly decodable from activations and, in turn, controllable via interventions along those directions.

### 5.1 Script Separability in Embedding Space

Following the linear representation hypothesis (Park et al., 2024), linearly separable classes occupy distinguishable regions of hidden-state space. We test whether this holds for script variants of the same language using three complementary measures: linear probes (Belinkov, 2022), layer-wise cosine similarity, and the Fisher discriminant ratio (Fisher, 1936) applied to paired native and romanized variants of identical sentences. Specifically, for each layer $\ell$ and input sentence $x$, we extract the mean-pooled hidden state $h_\ell(x) \in \mathbb{R}^d$ from the layer output and compute class means $\bar{h}_\ell^{\text{nat}}$ and $\bar{h}_\ell^{\text{lat}}$ over native and romanized FLORES sentences. All three analyses operate on these layer-wise representations.

Figure 4 shows consistent trends across all measures. Probe accuracy rises from near-chance to near-perfect in later layers, cosine similarity between native and romanized representations decreases with depth, and the Fisher ratio peaks in the same region. These results indicate that script

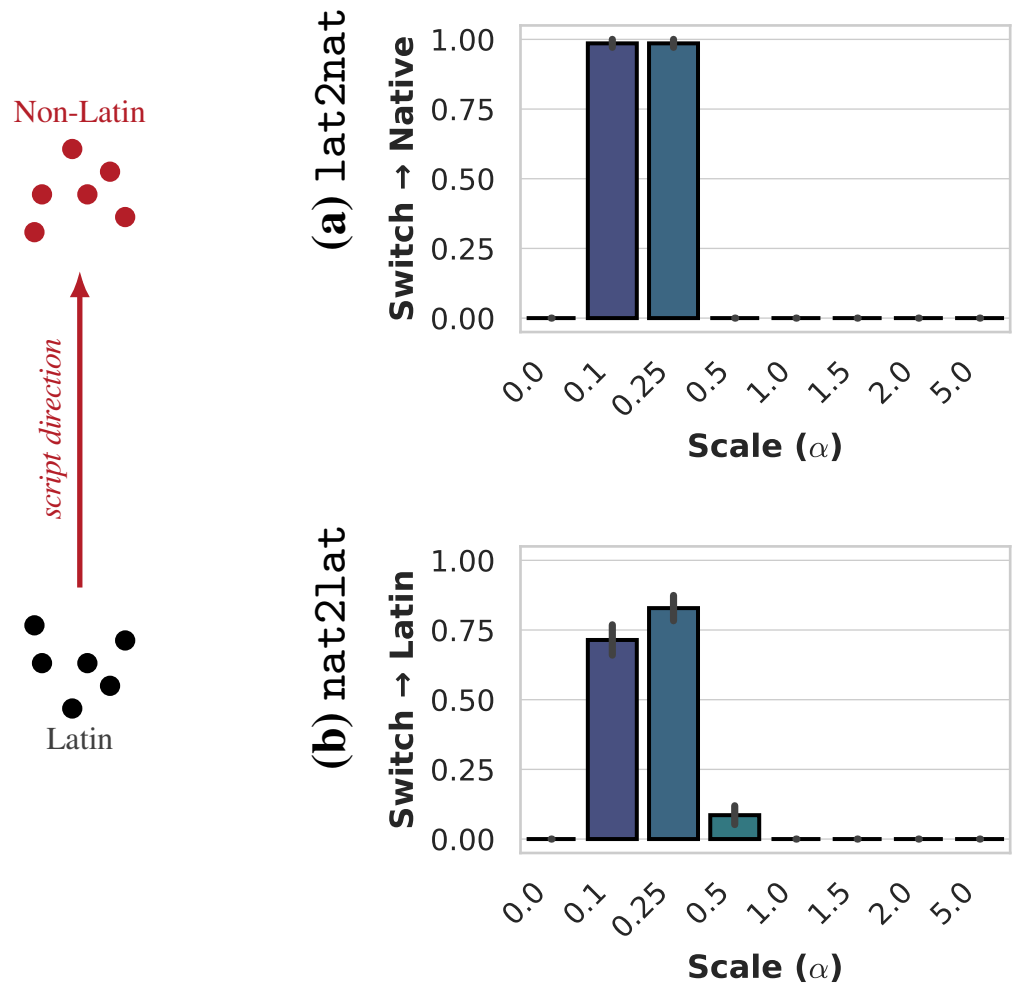

(c) Example

**Script Switching via Steering**

**Input (Devanagari):** मैं अपने समय प्रबंधन कौशल को कैसे सुधार सकता हूँ? *(How can I improve my time management skills?)*

**Steered Output (Latin Script):** Aapke samay prabandhan kaushal ko sudharne ke liye yahaan kuch sujhav hain: 1. Lakshya Nirdharit Karein: Apne din ke lakshyon ko likhein aur... *(Here are some tips to improve your time management skills: 1. Note: Write down your day's goals and...)*

Figure 5: Script switching success rate under all-layer steering for `lat2nat` (a) and `nat2lat` (b) in `Llama-3.1-8B`, alongside an example (c) where the Devanagari input is steered to produce output in Latin script. Arabic exhibits similar steering behavior, and we report the corresponding results in Figure 16 (Appendix B.2).

information becomes increasingly separable across layers, particularly in upper layers.

### 5.2 Script Switching via Representation Steering

Given the observed separability, we test whether script identity can be manipulated through linear interventions. For each layer, we compute the unit-normalized difference between the mean hidden representations of the two script classes: $v_\ell^{\texttt{nat2lat}} = \frac{\bar{h}_\ell^{\text{lat}} - \bar{h}_\ell^{\text{nat}}}{\|\bar{h}_\ell^{\text{lat}} - \bar{h}_\ell^{\text{nat}}\|}$. At inference time, we add $\alpha \cdot v_\ell$ to the residual stream of every layer simultaneously (Wang et al., 2025), using each layer's own vector rather than a shared vector, where $\alpha \in \mathbb{R}$ controls intervention strength. The reverse direction is defined as $v_\ell^{\texttt{lat2nat}} = -v_\ell^{\texttt{nat2lat}}$. We evaluate script switching using GlotLID (Kargaran et al., 2023), counting generations as successful when the predicted script matches the intervention target.

Figure 5(a,b) shows script-switching rates as a function of intervention strength under greedy decoding with 60-token generation. Moderate intervention strengths (e.g., $\alpha = 0.25$) achieve high switching rates in both directions, indicating that script identity is controllable through a linear intervention. The two directions are not equally effective, however: `lat2nat` reaches near-perfect switching at lower intervention strengths, whereas `nat2lat` is less reliable across the same range, an asymmetry that foreshadows the mechanistic findings in Section 6. The example in Figure 5(c) shows that steering changes surface script while largely preserving semantic content; we qualify semantic preservation in Appendix B.3.

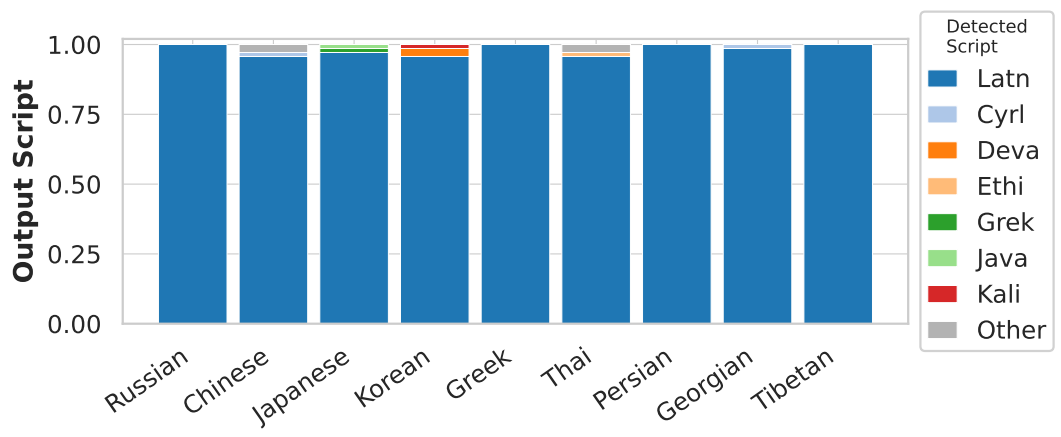

(a) `nat2lat`

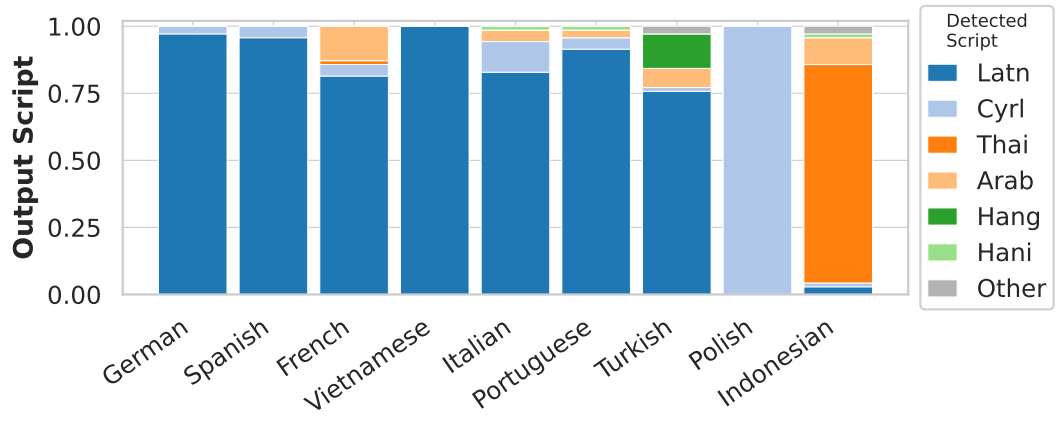

(b) `lat2nat`

Figure 6: Distribution of detected output scripts when steering vectors derived from Hindi and Arabic are applied to other languages in `Llama-3.1-8B`. For `nat2lat`, we test on Latin-script sources; for `lat2nat`, on Latin-script sources.

### 5.3 Generalization of Steering Directions to Other Languages

The steering vectors are derived using only Hindi and Arabic. To test whether they encode a general script axis rather than language-specific features, we construct a single `nat2lat` vector per layer by averaging the Hindi and Arabic vectors and do the same for the `lat2nat` vector. We apply each to prompts in languages unseen during vector construction and measure changes in output script distributions (Figure 6).[4]

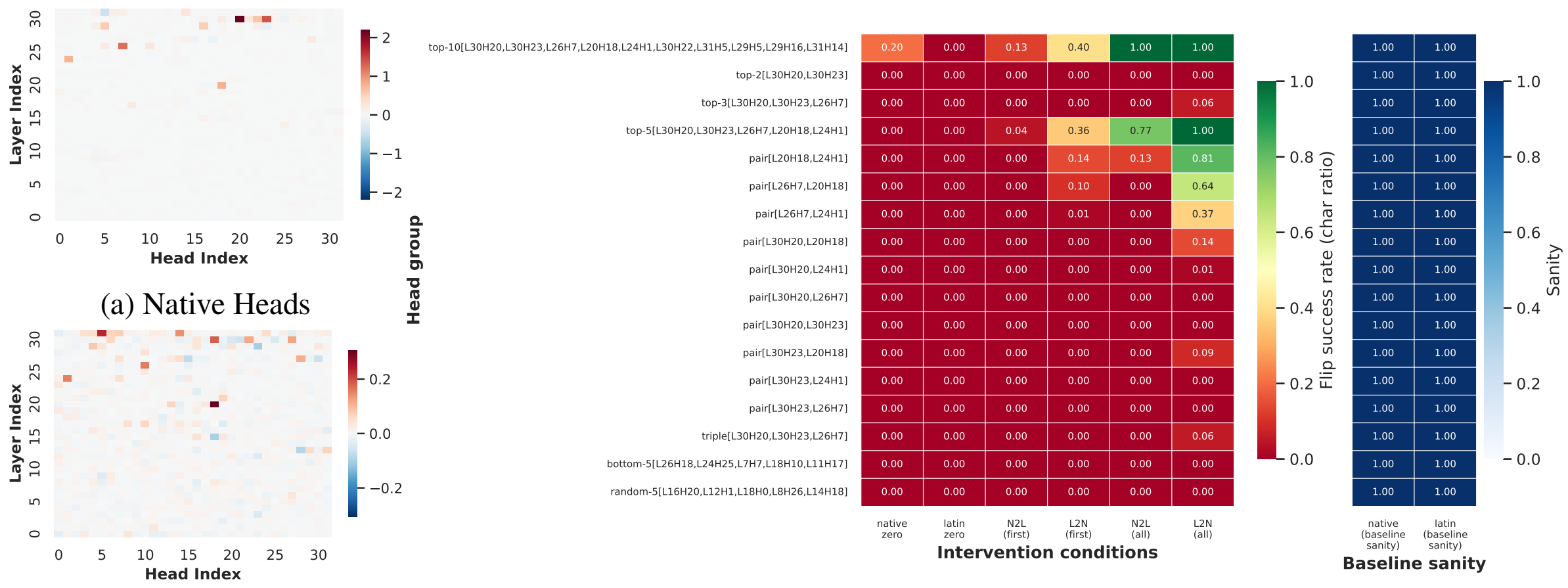

(a) Native Heads

(b) Latin Heads

(c) Top-$k$ Subset Validation

Figure 7: Mechanistic localization and validation of script-mediating heads in `Llama-3.1-8B` (Hindi). **(a, b)** Per-head causal contributions to script choice. Patching native activations into Latin-script runs (a) reveals a small set of late-layer heads (L30H20/23, L26H7, L20H18, L24H1) that carry native-script information; the reverse direction (b) shows diffuse, low-magnitude effects, with Latin emerging from many weakly contributing heads. **(c)** Only the top-5 subset is sufficient to flip generation from Latin to Devanagari (0.36 at first-token patching, 1.00 at sustained patching); smaller subsets, random, and bottom-5 controls fail. L20H18 plays a particularly important role. The baseline sanity columns (right of panel c) confirm that, without intervention, the model produces the expected script on each input. See Figures 24 and 28 (Appendices C.1 and C.2) for Arabic.

The directions behave differently. In most cases, the `nat2lat` direction converts outputs from a wide range of unseen non-Latin scripts, including Cyrillic, CJK, Greek, Thai, and Perso-Arabic systems, into Latin script. In contrast, `lat2nat` does not map Latin-script languages into a single target script. Instead, outputs occasionally drift toward varied non-Latin scripts that appear loosely compatible with the target language: Polish toward Cyrillic, Indonesian toward Thai, and French toward Arabic.

This suggests that the learned direction encodes a coarse "Latin versus non-Latin" distinction rather than a language-specific transliteration mapping. More broadly, the transfer results indicate that script information is represented along a shared direction that generalizes across writing systems unseen during steering vector construction. We note, however, that this direction is derived from only two script pairs; averaging over a broader set language-script pairs could give a more general direction, and we leave this to future work.

[4]As GlotLID does not support transliterated variants for all languages, scripts here are identified with Unicode ranges.

## 6 Mechanistic Analysis

Representation-level steering shows *that* script identity is encoded in the residual stream, but not *where* this information is computed. We therefore turn to attention heads as the unit of mechanistic localization.[5] As each head contributes additively to the residual stream through the output projection (Elhage et al., 2021), individual heads can be independently patched and causally evaluated.

### 6.1 Localizing Script-Mediating Heads

We use counterfactual activation patching (Meng et al., 2022; Vig et al., 2020) on 50 parallel native/romanized prompt pairs from FLORES. For each pair $(x_i^{\text{nat}}, x_i^{\text{lat}})$, we run the model twice and cache the pre-`o_proj` activation for every attention head $(\ell, h)$ at the last token position. We then perform a patched forward pass on a target prompt by replacing the cached activation slice for head $(\ell, h)$ with the corresponding activation from the source prompt. Let $z_t$ denote the next-token logit for vocabulary token $t$ after the patched forward pass. We measure the resulting change in script margin, $\Delta m = \log \sum_{t \in V_{\text{nat}}} e^{z_t} - \log \sum_{t \in V_{\text{lat}}} e^{z_t}$, where $V_{\text{nat}}$ and $V_{\text{lat}}$ denote vocabulary tokens beginning with native-script and Latin-script characters, respectively. $\Delta m$ quantifies the head's causal contribution to the log-odds of producing a native- vs. Latin-prefix token, with positive values indicating shifts toward native-script output.

[5]We do not rule out that MLP neurons, including the language-selective neurons identified by Tang et al. (2024) and Gurgurov et al. (2025), compute signals that these heads subsequently read out and see a full circuit-level account spanning attention and MLP components as an important direction for future work.

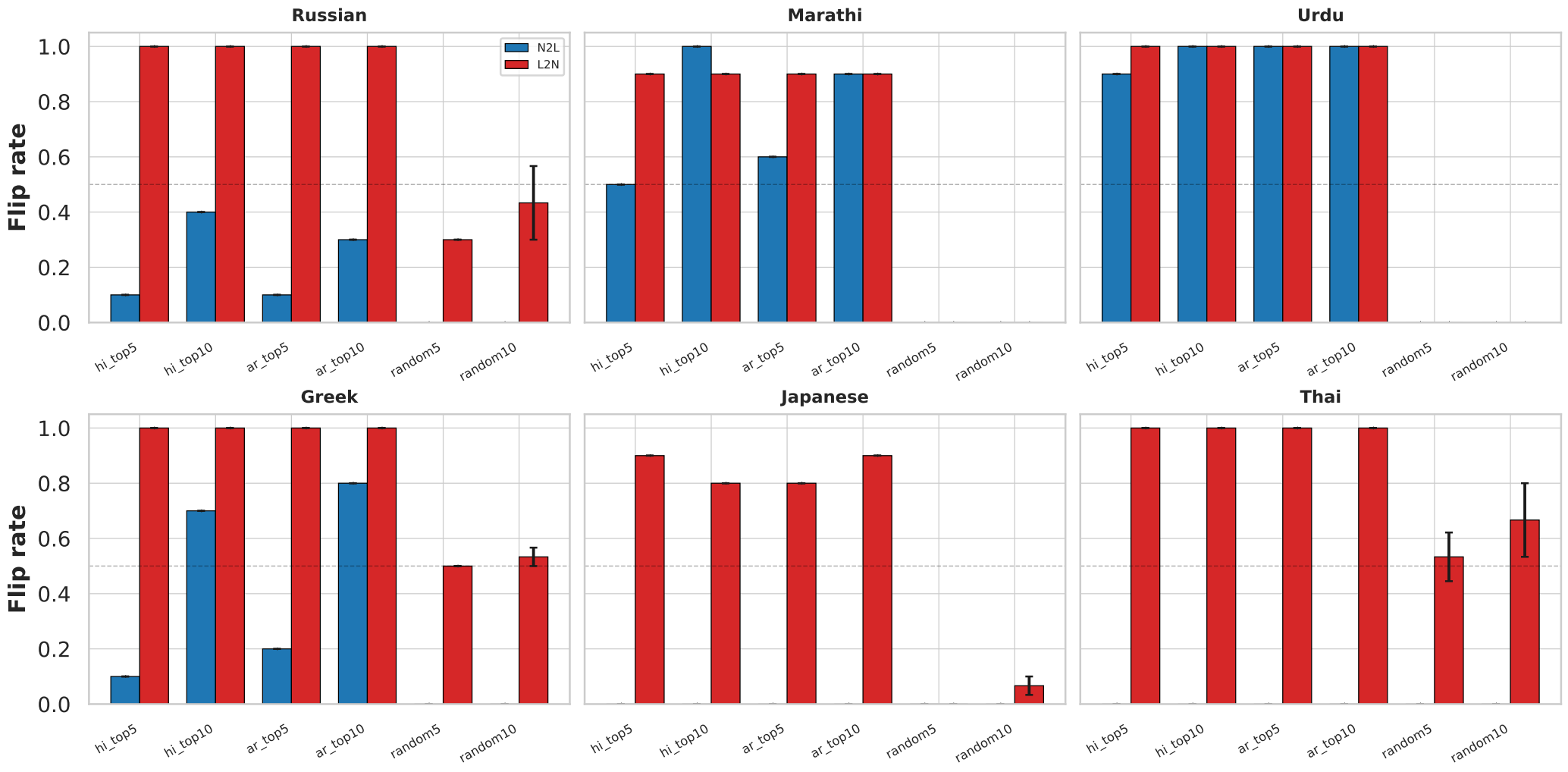

Figure 8: Cross-lingual transfer of script-mediating heads identified on Hindi and Arabic to five unseen languages in `Llama-3.1-8B` to additional languages. Sustained patching of the top head subsets flips output script for most languages. `L2N` (equivalent to `lat2nat`) patches native-script activation into Latin-script runs while, `N2L` denotes the opposite direction.

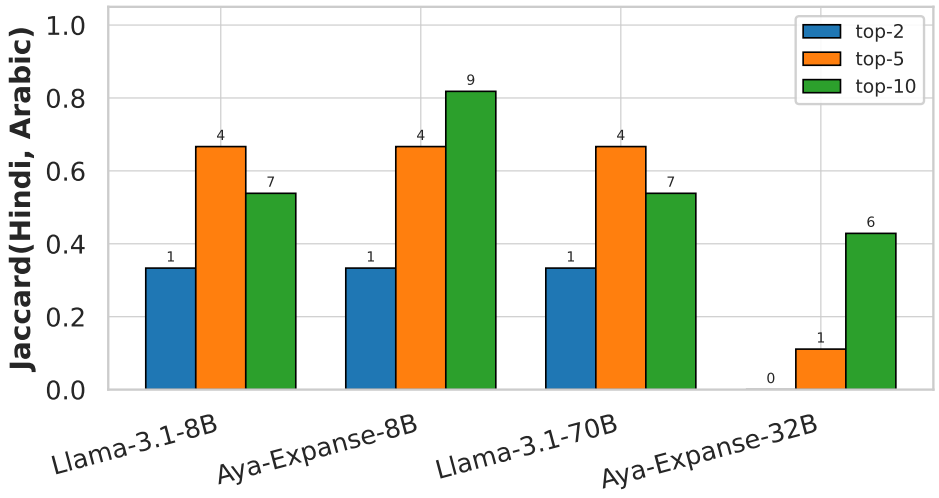

Figure 9: Jaccard overlap between top-k script-mediation heads identified on Hindi and on Arabic, for $k \in \{2, 5, 10\}$. Numbers above bars indicate the overlap.

We evaluate two patching directions using the same notation as in Section 5. `lat2nat` patches native activations into Latin runs, isolating heads that promote native-script generation. `nat2lat` performs the reverse intervention. Summing the $|\Delta m_{\text{nat2lat}}| + |\Delta m_{\text{lat2nat}}|$ ranks heads by overall script-mediating strength, regardless of which direction dominates.

Figure 7(a,b) reveals a strong asymmetry between the two directions. Native-script mediation is sharply localized: a small set of late-layer heads (L30H20, L30H23, L26H7, L20H18, and L24H1) produce large positive effects under `lat2nat` patching, while most heads remain near zero. Latin-script information is diffusely distributed: the `nat2lat` map has no dominant peaks, and effect magnitudes are much smaller (note the scales: $[-2, 2]$ vs. $[-0.3, 0.3]$).

This localization pattern mirrors the asymmetry observed in representation steering. Non-Latin generation depends on a compact set of identifiable components, whereas Latin-script generation appears to rely on distributed contributions across the model. We refer to this localized set of non-Latin-script generating heads as the *script-mediation gate*.

### 6.2 Validating the Gate via Interventions

To test whether the identified heads are causally sufficient for script control, we intervene during generation on 70 held-out prompts from CLaS-Bench. We generate 60-token continuations using greedy decoding and measure flip success rate.

We evaluate three intervention settings:

- **Zero ablation**: zero the target heads at every generation step.
- **Counterfactual patch (first token)**: replace head activations only at the first generated token using cached activations from the opposite-script run.
- **Counterfactual patch (all tokens)**: apply the replacement at every generation step.

| Lang. | Dir. | Prompt | Baseline | Patched (Top 10 heads) |
|---|---|---|---|---|
| **ru** | N2L | Напиши короткое стихотворение. *(Write a short poem.)* | Туман над рекой лежит, Золотые лучи солнца играют, Ветер дует, листья шумят... | Tymanyi mechta, kak zvezda v nochi, Golubaya, svetla, daleko v golubom. Ona svetit, oni svetit, bez granits... |
| | L2N | Napishi korotkoye stikhotvoreniye. *(romanized)* | Ya videl solntse vo snye, I pomnily o tebe. Mne stalo teper' leto, I ya zhdu tebya... | В темноте ночи я слышу звук, Как сердце бьется, как стук. В глубине души я чувствую пульс, И знаю, что я живу... |

Table 1: Patching the top-10 heads identified on Arabic causally flips the output script, even though Russian (Cyrillic) is unrelated to either training language's non-Latin script. Baseline outputs match the input script; patched outputs flip to the opposite script while preserving coherent content. More examples are in Appendix C.4.

We sweep multiple head subsets (top-2, top-3, top-5, top-10, and controls) to identify the minimal sufficient intervention set.

Figure 7(c) shows that top-5 patching achieves full success under sustained intervention: the model flips from Latin to Devanagari on 100% of prompts when heads are patched throughout generation, but only 36% under first-token patching alone. Sustained intervention is required, and the gap between conditions reveals that the gate operates continuously across generation rather than at a single decision point. All smaller subsets (top-2, top-3, every pair within top-5) mostly fail at 0%. The jump from 0% at top-3 to 100% at top-5 indicates a compact but redundant gate: no individual pair carries the effect, and at least two of {L20H18, L24H1} must accompany the two layer-30 heads and L26H7. Furthermore, Random-5 and Bottom-5 head sets achieve 0% flips in every condition.

Behavioral flip rates are higher in the `lat2nat` direction (0.36/1.00 at first/all) than in `nat2lat` (0.04/0.77). Patching native-writers into a Latin run injects a strong signal that overrides the default; patching them out of a native run removes the signal but does not actively insert Latin content, consistent with what the localization heatmap from Figure 7b predicts. Expanding to the top-10 heads narrows the gap, consistent with the diffuse-Latin picture: more heads are needed to assemble enough Latin signal to flip a native baseline.

Zeroing the top-10 heads on native input flips 20% of generations to Latin (0% in the reverse direction). When the gate is silenced, the network's default pathways take over for a fifth of the prompts, defaulting to Latin output, consistent with Latin being the model's fallback substrate. The remaining 80% still produce native script, indicating that other pathways can compensate for the silenced heads.

### 6.3 Gate Transfer across Languages and Scripts

If the identified heads encode language-specific content, they should not transfer across unrelated languages and writing systems. We first test this by comparing the top-$k$ heads identified independently on Hindi and Arabic. Figure 9 shows high overlap between the two sets across all models, particularly at larger values of $k$. This suggests that the same heads participate in script mediation across languages with unrelated non-Latin scripts.

We next evaluate transfer on five unseen languages spanning four writing systems: Russian (Cyrillic), Marathi (Devanagari), Urdu (Perso-Arabic), Greek (Greek), and Japanese (Kanji, Hiragana, and Katakana). The heads are identified using Hindi or Arabic, but activations are patched using native/romanized pairs from each target language itself. We identify the scripts using their unicode markers.

Figure 8 shows that the identified head sets transfer across most evaluated languages under sustained patching, while random-head controls remain near chance. Table 1 illustrates a representative example in Russian: patching heads identified on Arabic successfully flips the output script in languages whose writing systems were not used during localization. The `nat2lat` direction transfers less reliably across all tested languages and benefits from larger head sets, consistent with the localization asymmetry from Section 6.1.

Taken together, these findings suggest that the identified heads implement a language-agnostic mechanism for script mediation rather than a language-specific one. Across localization (§6.1), intervention (§6.2), and transfer experiments (§6.3), non-Latin generation consistently depends on compact, identifiable components, while Latin-script generation remains more diffusely distributed across the network.

## 7 Conclusion

In this work, we investigated how LLMs internally handle languages written in multiple scripts. Using logit lens analysis during transliteration, we showed that models systematically perform latent romanization, using the Latin script as an intermediate representation bridge. At the representational level, we found that scripts of the same language become progressively separable across layers and that a simple linear direction reliably flips the model's output script while preserving meaning, transferring even to languages and writing systems unseen during construction. At the mechanistic level, we localized a small set of late-layer attention heads that causally mediate script choice, a script-mediation gate that is sufficient to redirect output, language-agnostic, and transferable across writing systems. The same asymmetry recurs at every level: non-Latin output is produced by a compact identifiable mechanism, while Latin emerges from diffuse contributions throughout the network, consistent with Latin acting as the model's default output substrate.

To conclude, these results support the central hypothesis introduced in Section 1: *Latin script appears to function as a privileged substrate for surface-language representations in LLMs.*

## Limitations

Our analysis focuses on four open instruction-tuned models from two families. While we observe consistent patterns across most models, the gate's exact composition (which specific layer-head pairs participate) varies across models, and we cannot rule out that closed or differently-trained models would exhibit qualitatively different mechanisms. We also restrict steering vector construction and head identification to two primary languages, Hindi and Arabic; while we test transfer to additional languages, all of them have established romanization conventions and substantial pretraining presence. Languages whose romanizations are rarely encoded in pretraining data may behave differently. Finally, we do not investigate whether the gate is part of a larger underlying circuit, which would require substantially more computational resources and is left for future work.

## Acknowledgments

This research was supported by the German Federal Ministry of Research, Technology and Space (BMFTR) as part of the project TRAILS (01IW24005).

Katharina Trinley's work was funded by the Deutsche Forschungsgemeinschaft (DFG, German Research Foundation) - Project number 579327477.

# Appendix

## A Per-Layer Output Analysis: Additional Details and Results

### A.1 Logit lens

Generally, in a decoder only LLM, the unembedding matrix is multiplied with the final hidden state and a softmax is taken on the product to produce the token distributions at that token generation step. Since all hidden states of an LLM are in the same shape, it is possible to apply the unembedding matrix and softmax on all layers, thereby generating token distributions at all layers. This method of prematurely decoding hidden states is referred to as *logit lens* (Nostalgebraist, 2020). Logit lens reveals how the latent representations evolve across layers to produce the final output, providing insights into the progression of computations within the model. Because intermediate hidden states are not trained to be read out by the final unembedding matrix, logit lens readouts can be poorly calibrated, especially at early layers; we restrict the analysis to the last ten layers where romanized subwords are visually legible in qualitative examples and treat this as a decodability signal rather than direct causal evidence.

### A.2 Sample Prompt

A sample prompt to transliterate a word from Devanagari to Latin is shown below:

“मछली" - Latin - “machli"
“आम" - Latin - “aam
“भाई" - Latin - “bhai"
“गंध" - Latin - “gandh"
“सूरज" - Latin - “suraj"
“फूल" - Latin -

### A.3 Latent Fraction

Formally, we compute the latent fraction as follows:

For layer $l$, timestep $t$, sample $i$ and set of corresponding romanized tokens $R$ :
1. **Latent romanization condition:**

$$r_{l,t}^{(i)} = \begin{cases} 1, & \text{if } \max_{r \in R} P(x_t = r \mid l, t) > 0.1 \\ 0, & \text{otherwise} \end{cases}$$

2. **Latent fraction for a layer $\ell$:**

$$\text{L.F}(l) = \frac{1}{N} \sum_{i=1}^{N} \frac{1}{T} \sum_{t=1}^{T} r_{l,t}^{(i)}$$

where $N$ is the number of samples, $T$ is the number of generation timesteps and $P(x_t = r|l, t)$ is the probability of generating token $r$ at timestep $t$ and layer $\ell$.

### A.4 Additional Results

Figure 10 shows per-layer output analysis for `Llama-3.1-70B`, `Llama-2-13B`, `Aya-Expanse-8B`, and `Aya-Expanse-32B` with Malayalam as the source language and Devanagari, Cyrillic, Arabic, Greek and Telugu as the target scripts. Figure 11 depicts per-layer output analysis for the models with Greek as the source language and Devanagari, Cyrillic, Arabic, Gujarati and Telugu as the target scripts. Figure 12 depicts qualitative logit lens analysis for transliteration of the Hindi word for “rain” from Devanagari to Cyrillic.

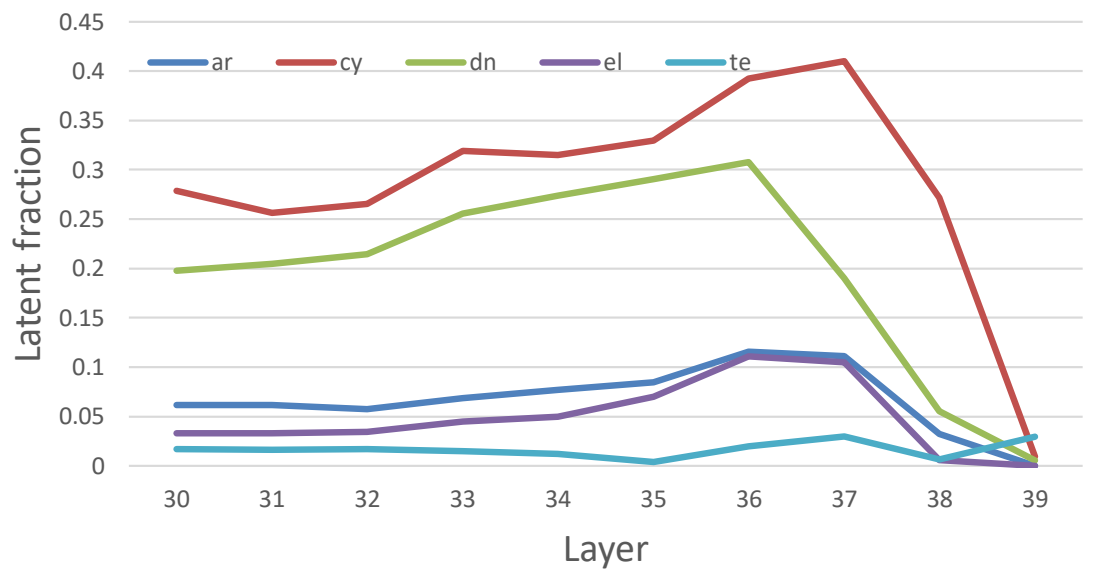

(a) Llama-2-13B

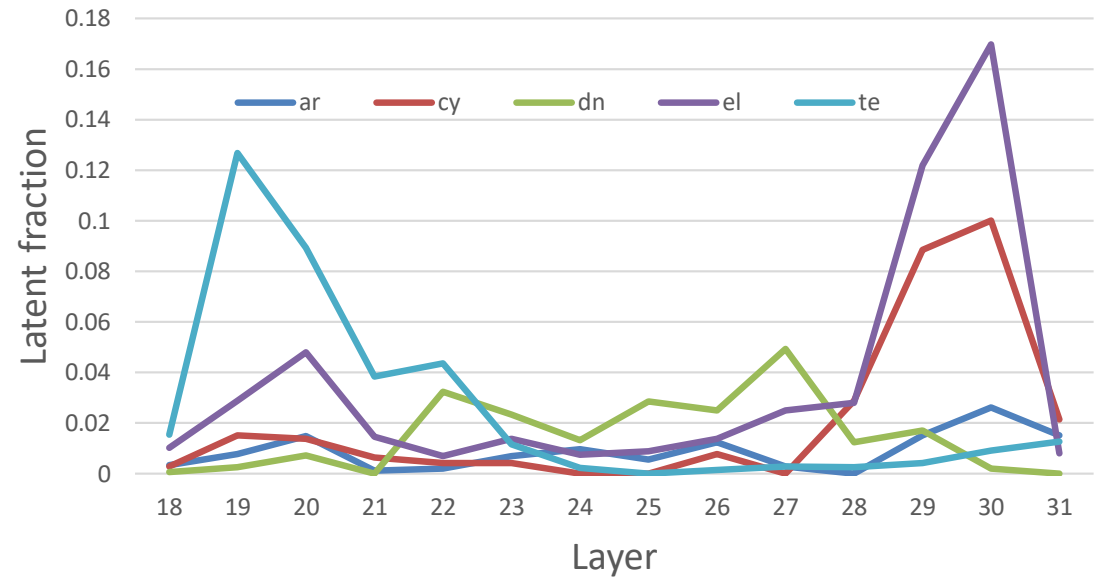

(b) Aya-Expanse-8B

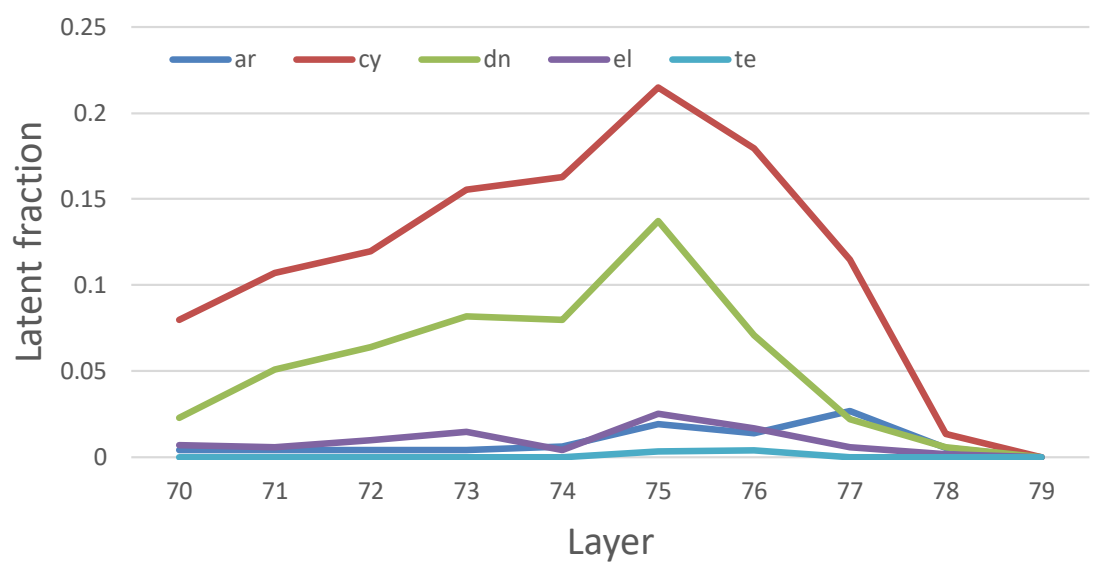

(c) Llama-3.1-70B

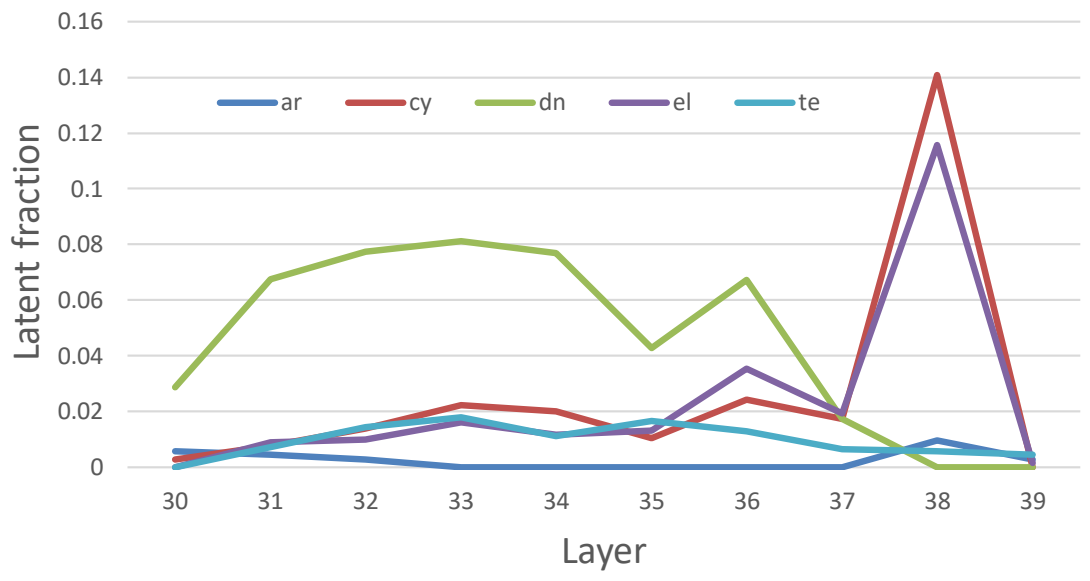

(d) Aya-Expanse-32B

Figure 10: Distribution of romanized tokens across model layers (plotted across the last 10 layers) for `Llama-3.1 8B`, `Llama-3.1 70B`, `Llama-2 13B`, `Aya Expanse-8B` and `Aya Expanse-32B` during transliteration with Malayalam as the source language, averaged across 130+ samples. $X$-axis represents layer index, $y$-axis represents latent fraction i.e. the fraction of timesteps where romanized tokens occur with a probability $> 0.1$ averaged over samples for a specific layer. We plot the distributions for 5 target Scripts: Arabic (ar), Telugu (te), Devanagari (hi), Cyrillic (cy) and Greek (el).

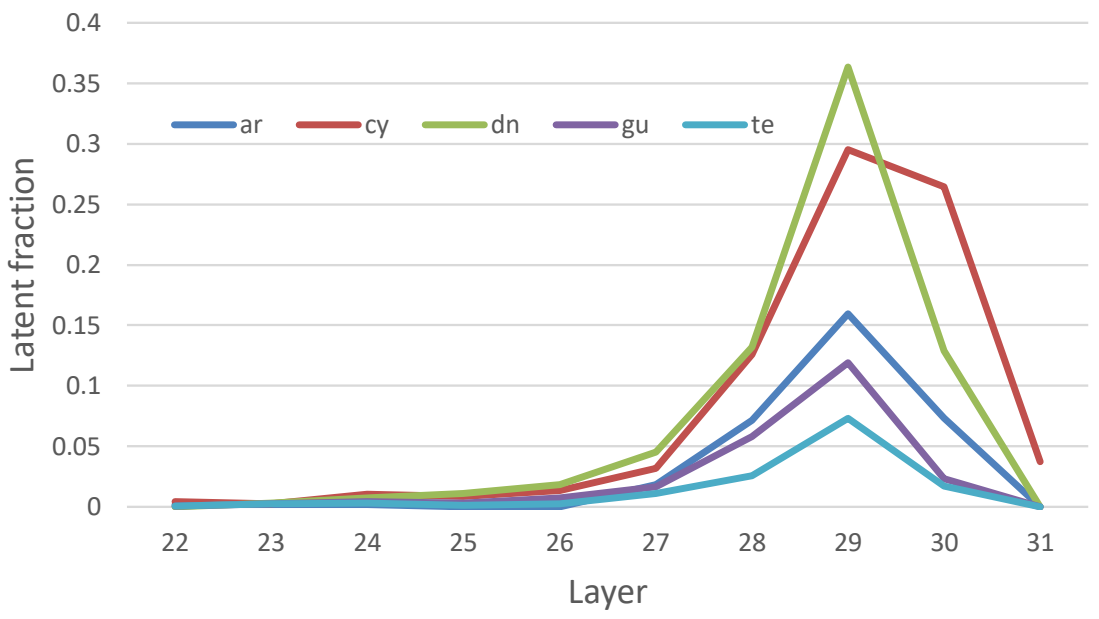

(a) Llama-3.1-8B

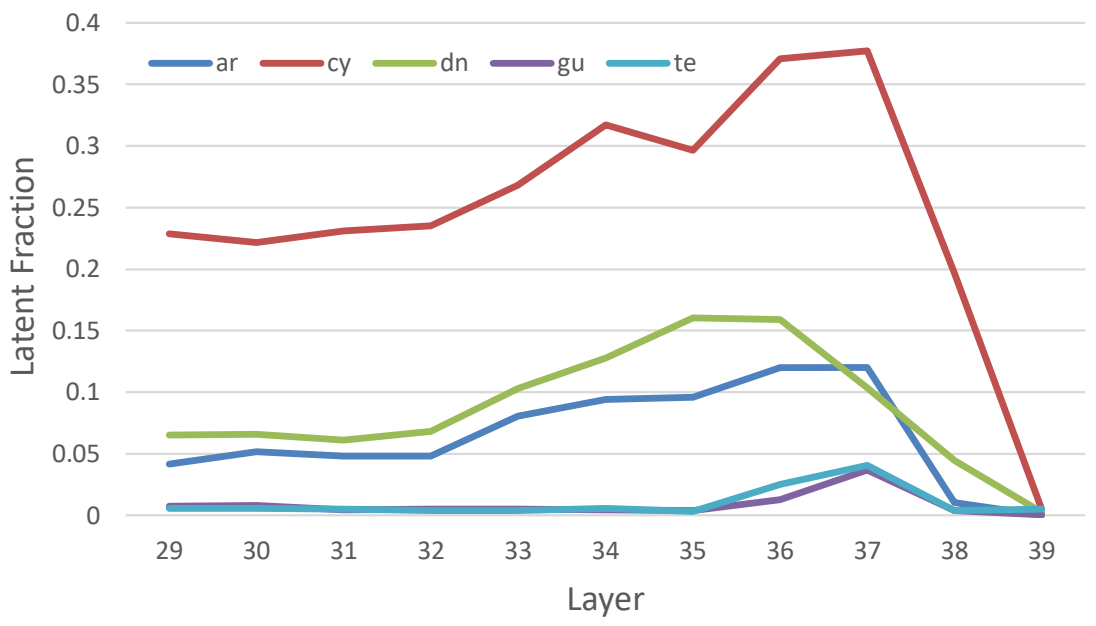

(b) Llama-2-13B

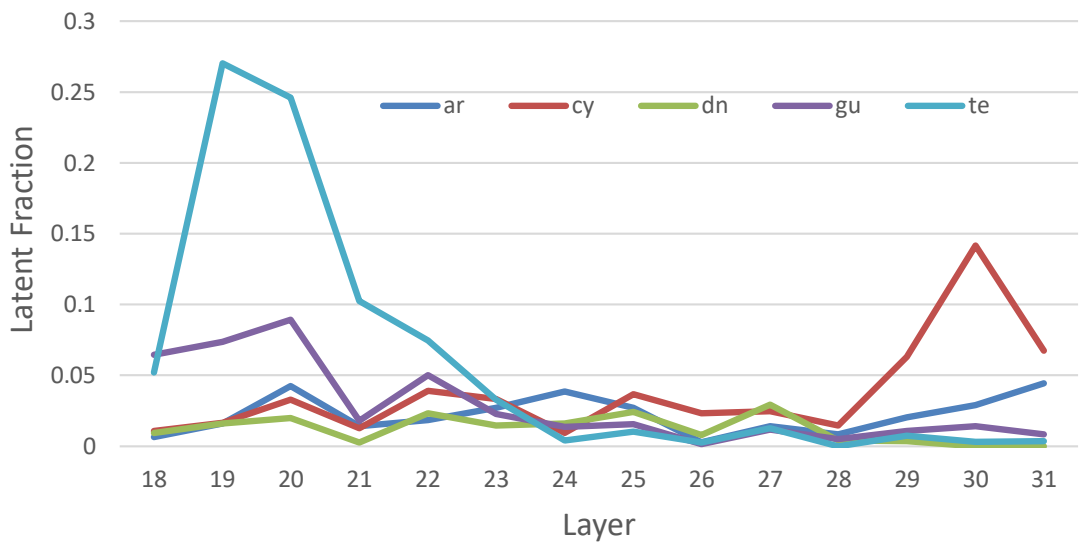

(c) Aya-Expanse-8B

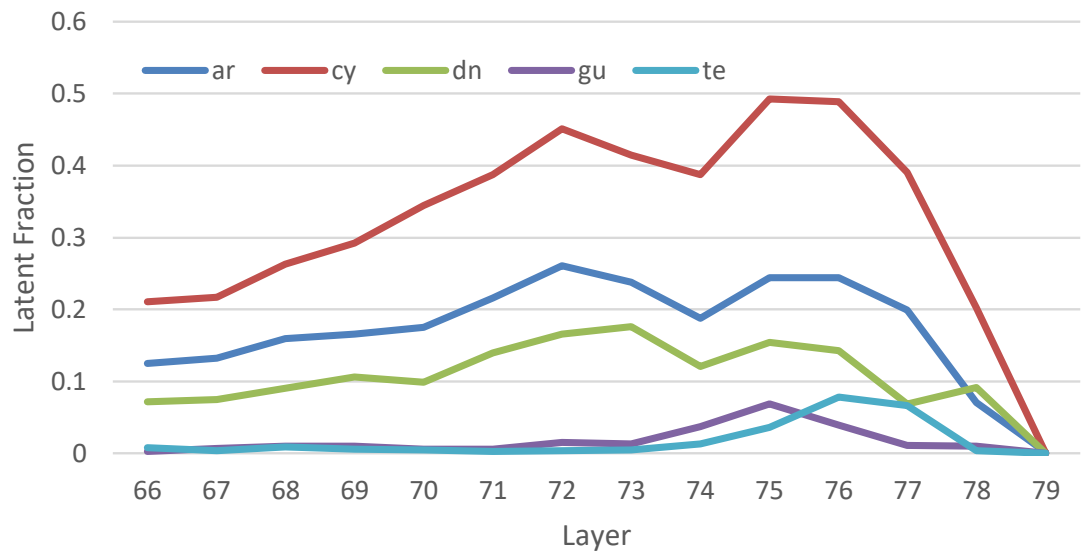

(d) Llama-3.1-70B

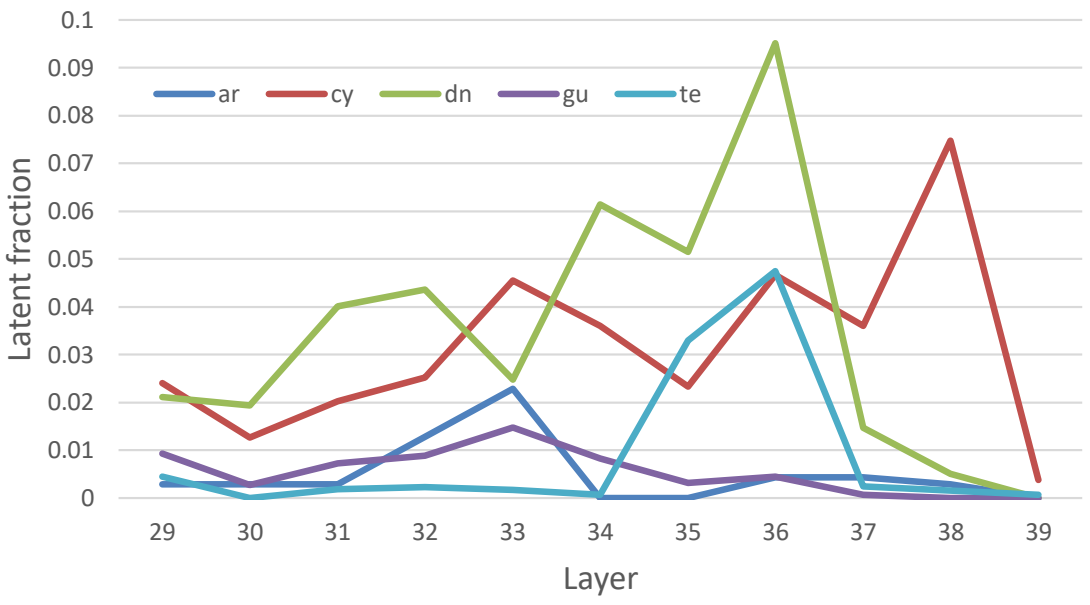

(e) Aya-Expanse-32B

Figure 11: Distribution of romanized tokens across model layers (plotted across the last 10 layers) for `Llama-3.1 8B`, `Llama-3.1 70B`, `Llama-2 13B`, `Aya Expanse-8B` and `Aya Expanse-32B` during transliteration with Greek as the source language, averaged across 130+ samples. $X$-axis represents layer index, $y$-axis represents latent fraction i.e. the fraction of timesteps where romanized tokens occur with a probability $> 0.1$ averaged over samples for a specific layer. We plot the distributions for 5 target Scripts: Arabic (ar), Telugu (te), Devanagari (hi), Cyrillic (cy) and Gujarati (gu).

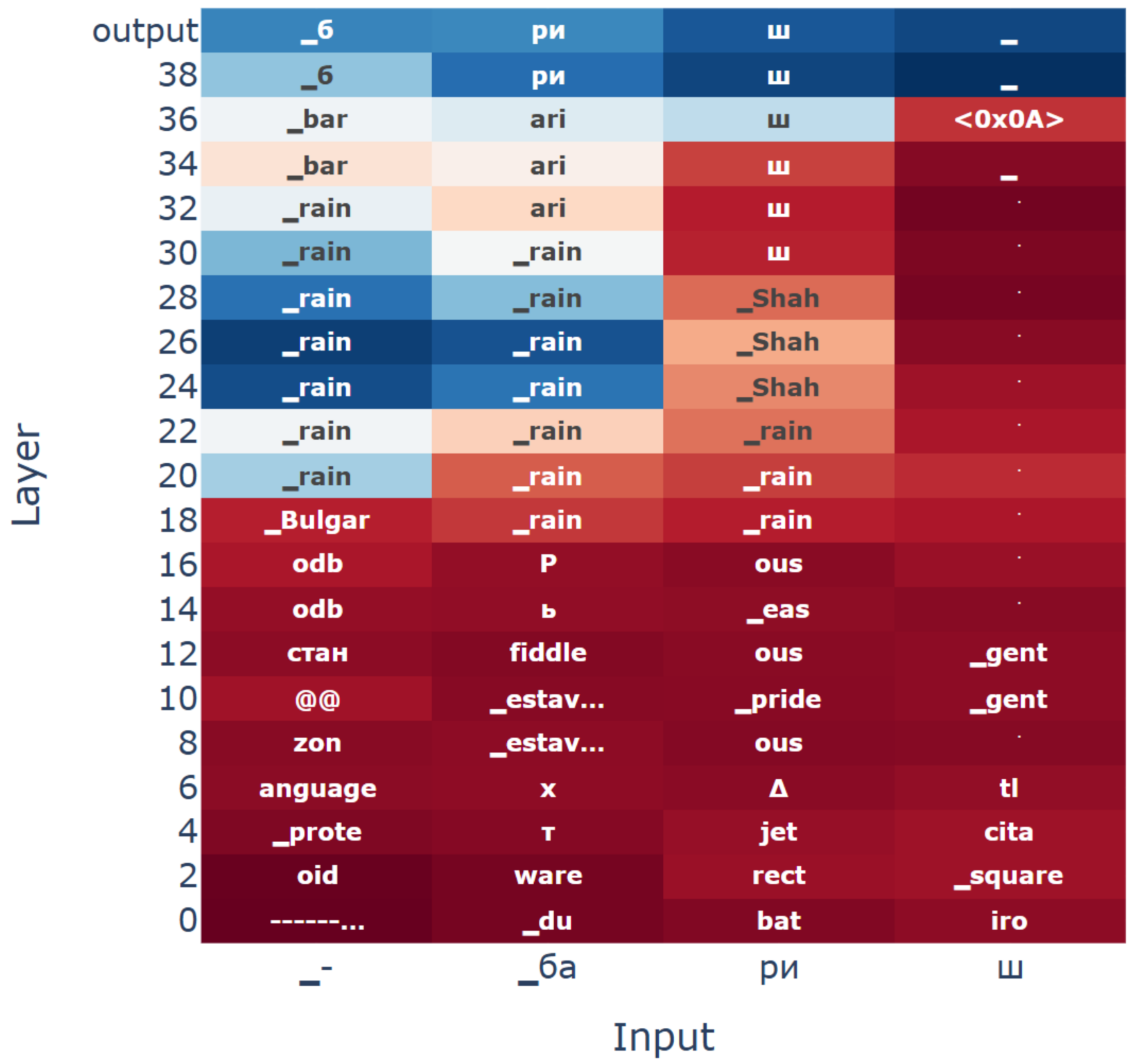

Figure 12: Logit lens visualization of Llama-2-13B transliterating the Hindi word for "rain" from Devanagari to Cyrillic. The plot shows the next-token distribution at each position (x-axis) across layers (y-axis), with the final output (бариш, "barish" in romanized form) taking shape from the bottom up. In the middle-to-upper layers, romanized subwords of the target word (бар – bar; ари – ari) appear before being transformed into the target script.

# B Representational Analysis

## B.1 Script Separability

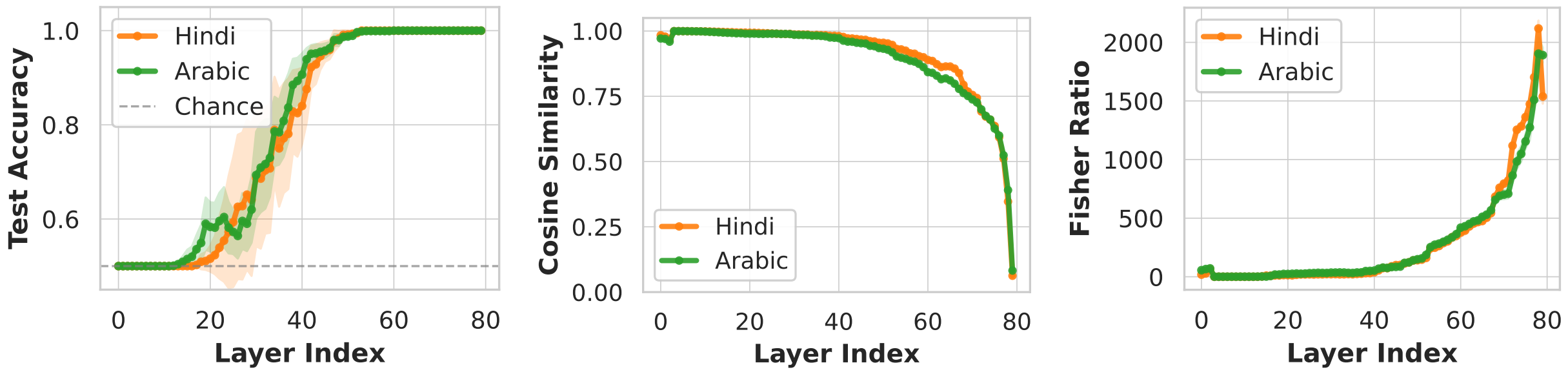

(a) Probe accuracy (b) Cos. similarity (c) Fisher ratio

Figure 13: Layer-wise probe accuracy, within-language similarity, and Fisher ratio for Hindi and Arabic. All metrics indicate that script separability emerges primarily in the later layers of `Llama-3.1-70B`.

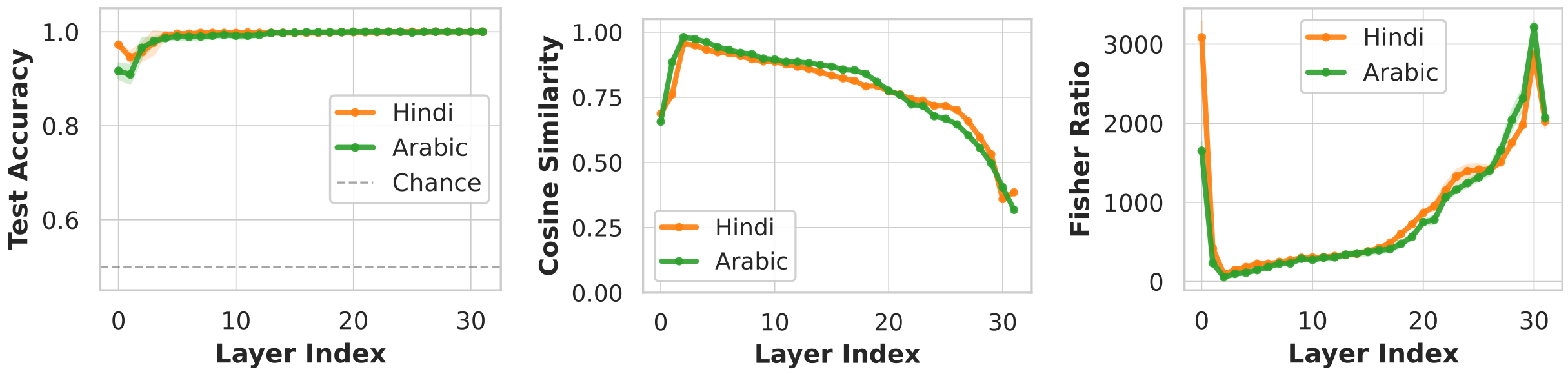

(a) Probe accuracy (b) Cos. similarity (c) Fisher ratio

Figure 14: Layer-wise probe accuracy, within-language similarity, and Fisher ratio for Hindi and Arabic. All metrics indicate that script separability emerges primarily in the later layers of `Aya-Expanse-8B`. Probe accuracy is high across most layers, likely reflecting overfitting given the model's stronger representational signal; cosine similarity and the Fisher ratio show separability emerging in the later layers, consistent with the `Llama` models.

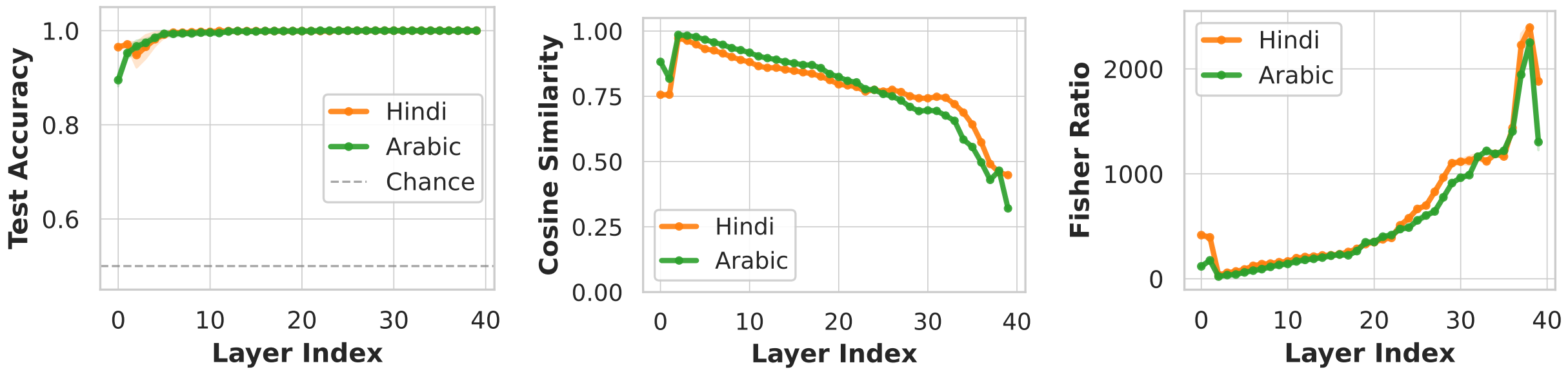

(a) Probe accuracy (b) Cos. similarity (c) Fisher ratio

Figure 15: Layer-wise probe accuracy, within-language similarity, and Fisher ratio for Hindi and Arabic. All metrics indicate that script separability emerges primarily in the later layers of `Aya-Expanse-32B`. Probe accuracy is high across most layers, likely reflecting overfitting given the model's stronger representational signal; cosine similarity and the Fisher ratio show separability emerging in the later layers, consistent with the `Llama` models.

## B.2 Representation Steering: Script Switching Rates

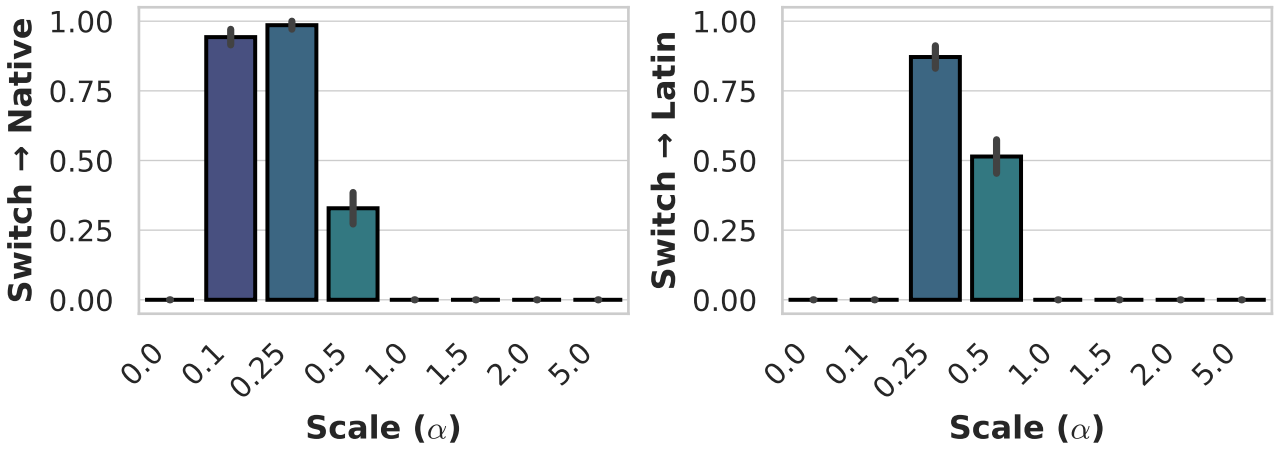

(a) Arabic (`lat2nat`) (b) Arabic (`nat2lat`)

Figure 16: Script switching results for Arabic in `Llama-3.1-8B`.

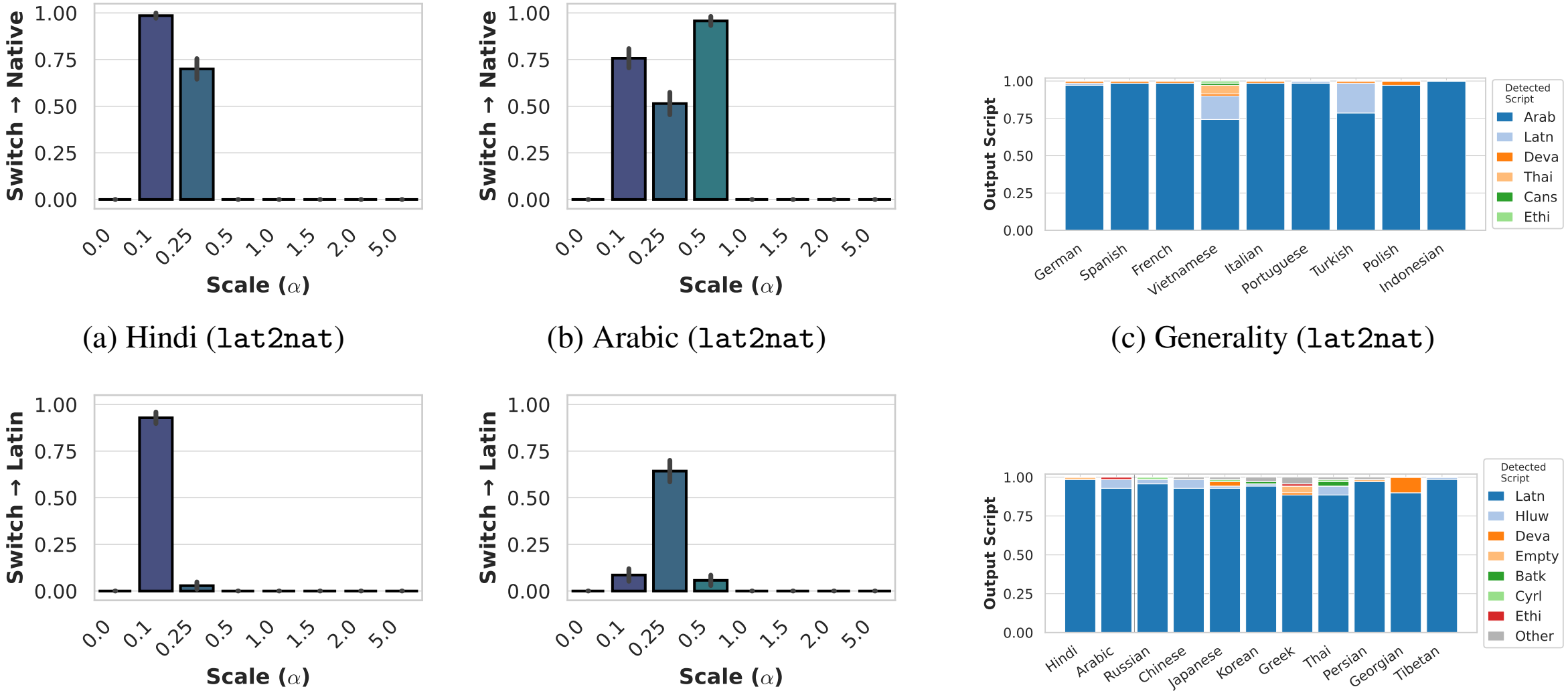

(a) Hindi (`lat2nat`) (b) Arabic (`lat2nat`) (c) Generality (`lat2nat`)

(d) Hindi (`nat2lat`) (e) Arabic (`nat2lat`) (f) Generality (`nat2lat`)

Figure 17: Script switching results across languages for `Llama-3.1-70B`. Left and middle columns show Hindi and Arabic steering performance across various alpha levels. The wider generality column highlights cross-lingual effects beyond the source languages.

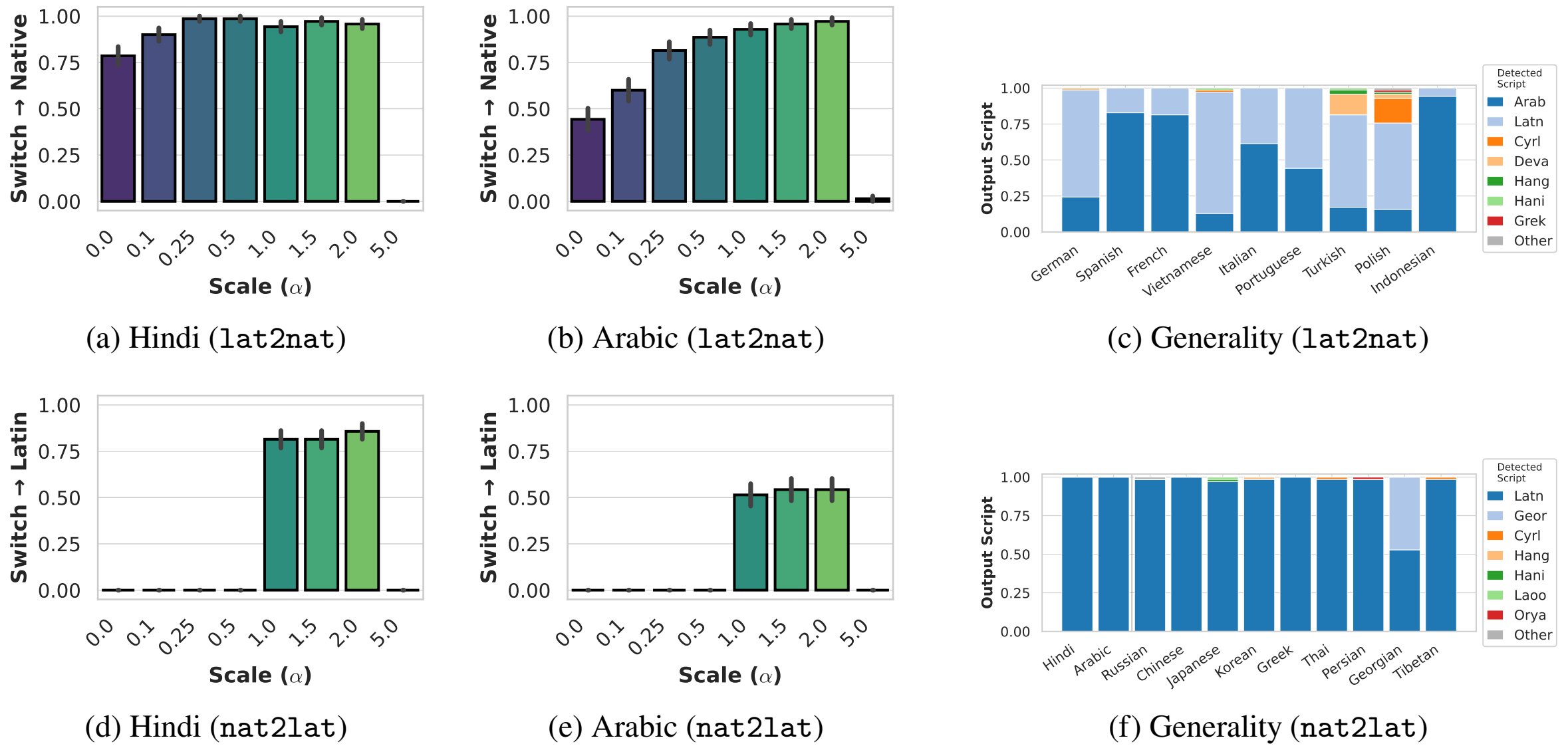

(a) Hindi (`lat2nat`) (b) Arabic (`lat2nat`) (c) Generality (`lat2nat`)

(d) Hindi (`nat2lat`) (e) Arabic (`nat2lat`) (f) Generality (`nat2lat`)

Figure 18: Script switching results across languages for **`Aya-Expanse-8B`**. Left and middle columns show Hindi and Arabic steering performance across various alpha levels. The wider generality column highlights cross-lingual effects beyond the source languages.

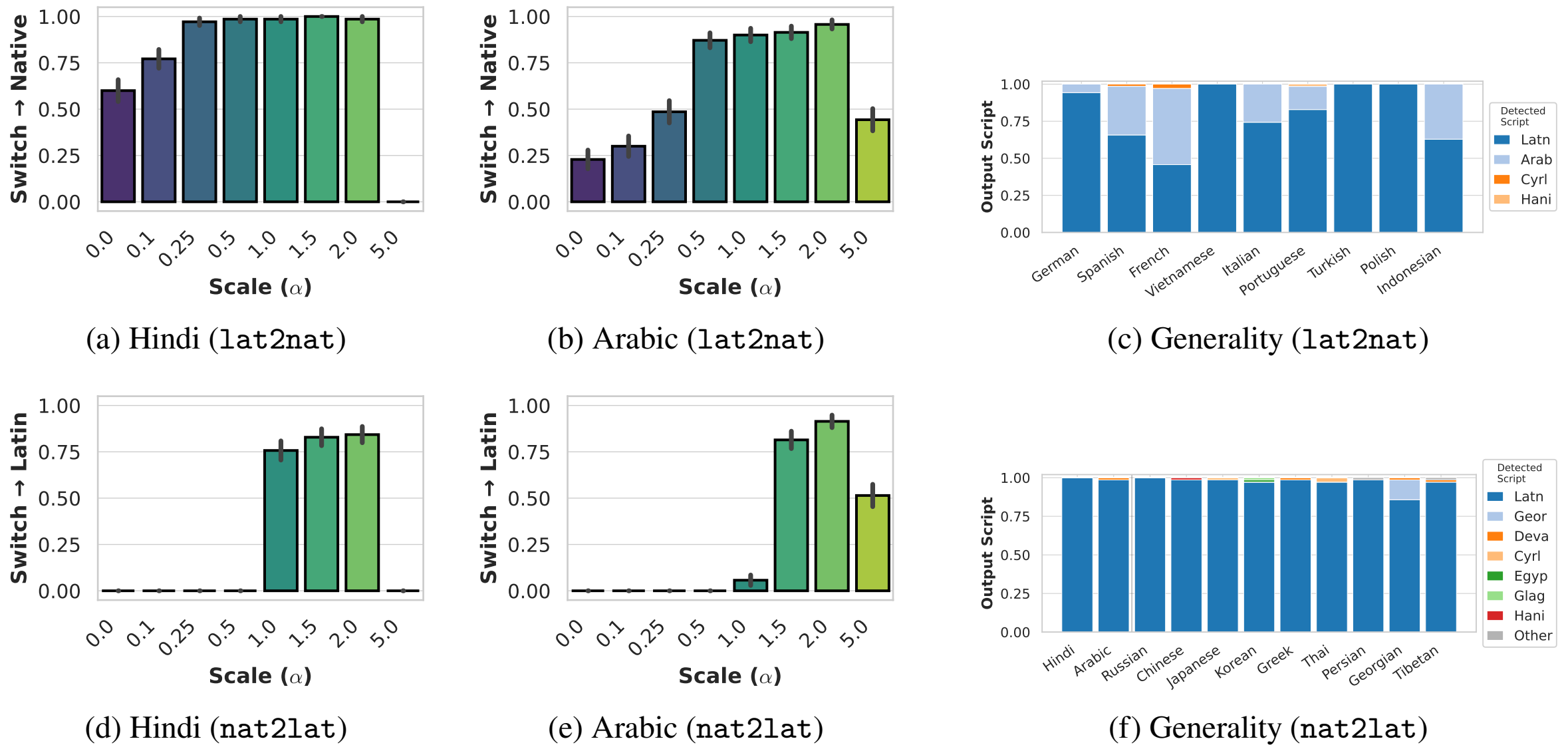

(a) Hindi (`lat2nat`) (b) Arabic (`lat2nat`) (c) Generality (`lat2nat`)

(d) Hindi (`nat2lat`) (e) Arabic (`nat2lat`) (f) Generality (`nat2lat`)

Figure 19: Script switching results across languages for **`Aya-Expanse-32B`**. Left and middle columns show Hindi and Arabic steering performance across various alpha levels. The wider generality column highlights cross-lingual effects beyond the source languages.

### B.3 Representation Steering: Output Quality Assessment

We assess response quality following the prompt-based evaluation protocol of Gurgurov et al. (2026), using Qwen3-8B (Yang et al., 2025) as an LLM judge with thinking mode disabled. The judge evaluates relevance and coherence on a 0-2 scale, with higher scores indicating more relevant, coherent, and useful responses.

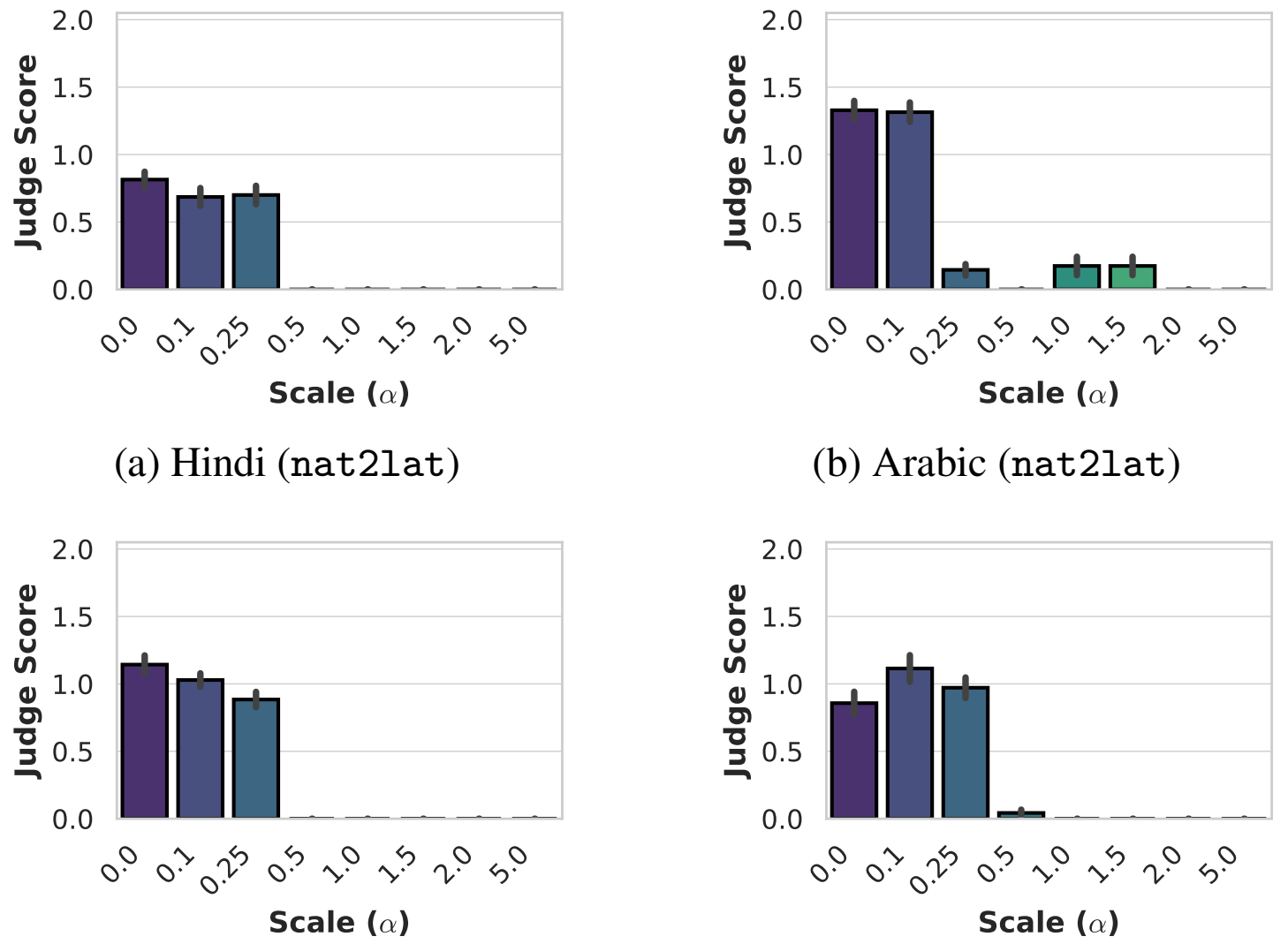

(a) Hindi (`nat2lat`) (b) Arabic (`nat2lat`)

(c) Hindi (`lat2nat`) (d) Arabic (`lat2nat`)

Figure 20: Response relevance after steering across languages for `Llama-3.1-8B`. Results show the quality of steered responses across different alpha levels for Hindi and Arabic in both steering directions.

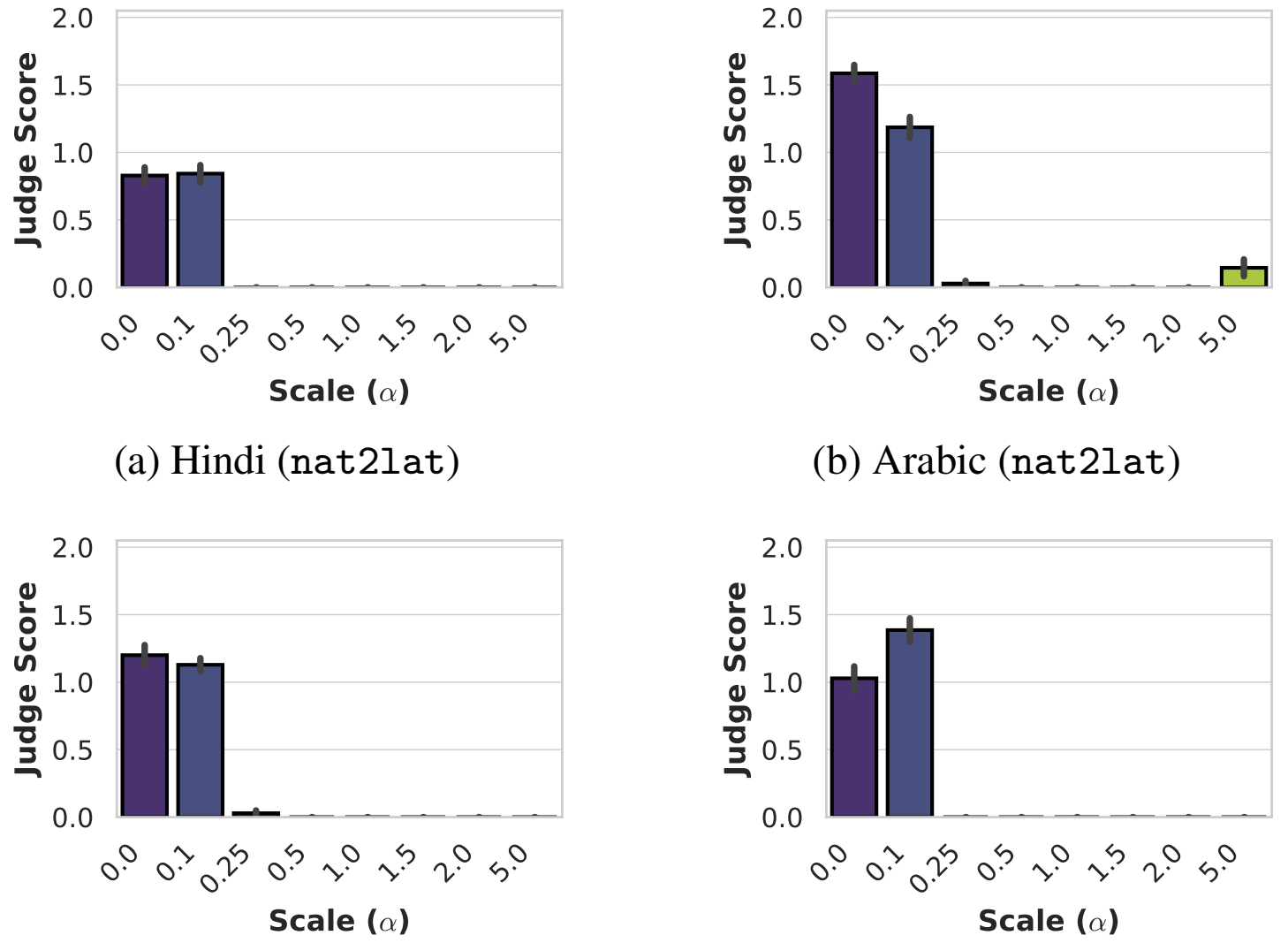

(a) Hindi (`nat2lat`) (b) Arabic (`nat2lat`)

(c) Hindi (`lat2nat`) (d) Arabic (`lat2nat`)

Figure 21: Response relevance after steering across languages for `Llama-3.1-70B`. Results show the quality of steered responses across different alpha levels for Hindi and Arabic in both steering directions.

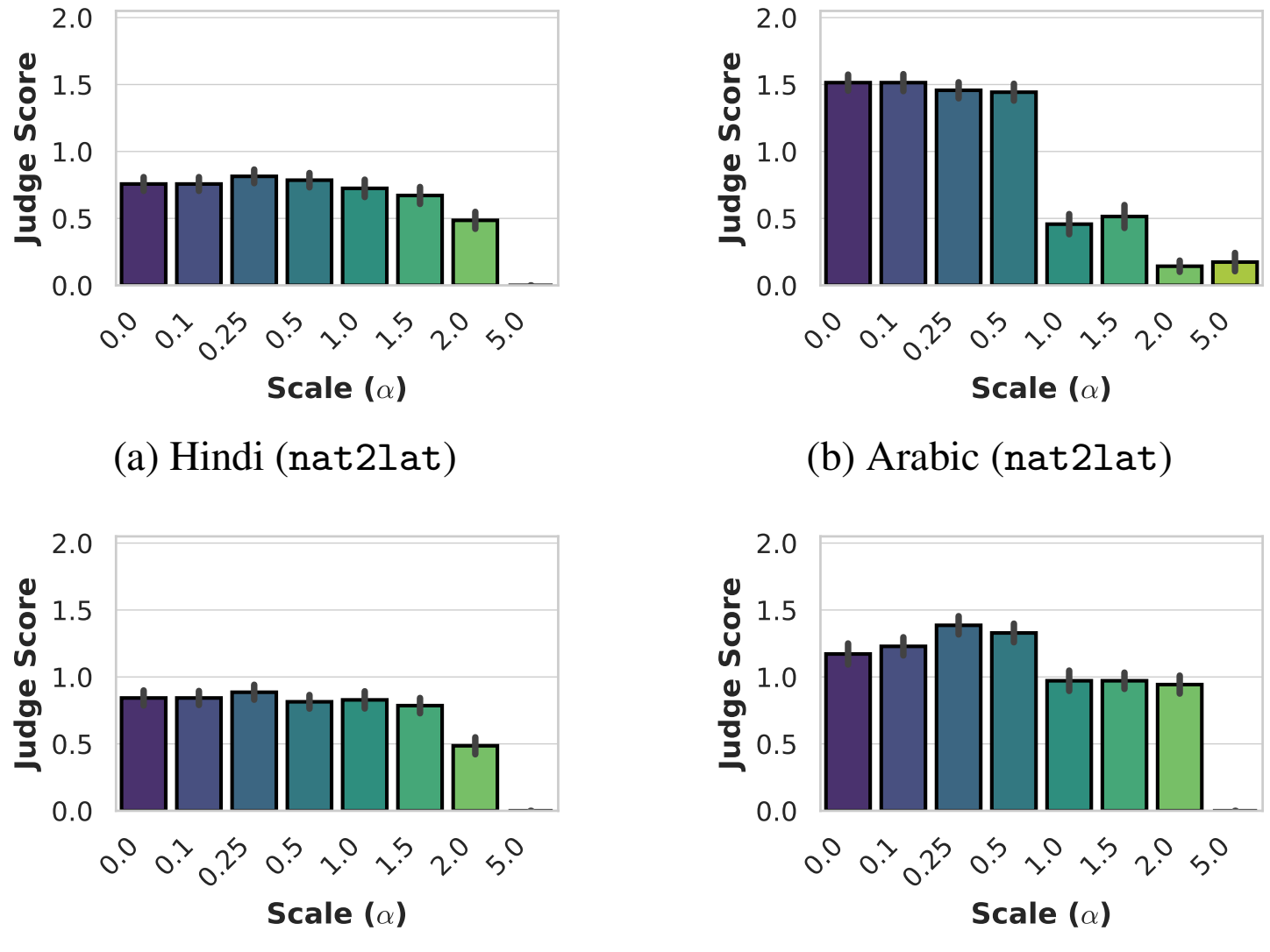

(a) Hindi (`nat2lat`) (b) Arabic (`nat2lat`)

(c) Hindi (`lat2nat`) (d) Arabic (`lat2nat`)

Figure 22: Response relevance after steering across languages for **`Aya-Expanse-7B`**. Results show the quality of steered responses across different alpha levels for Hindi and Arabic in both steering directions.

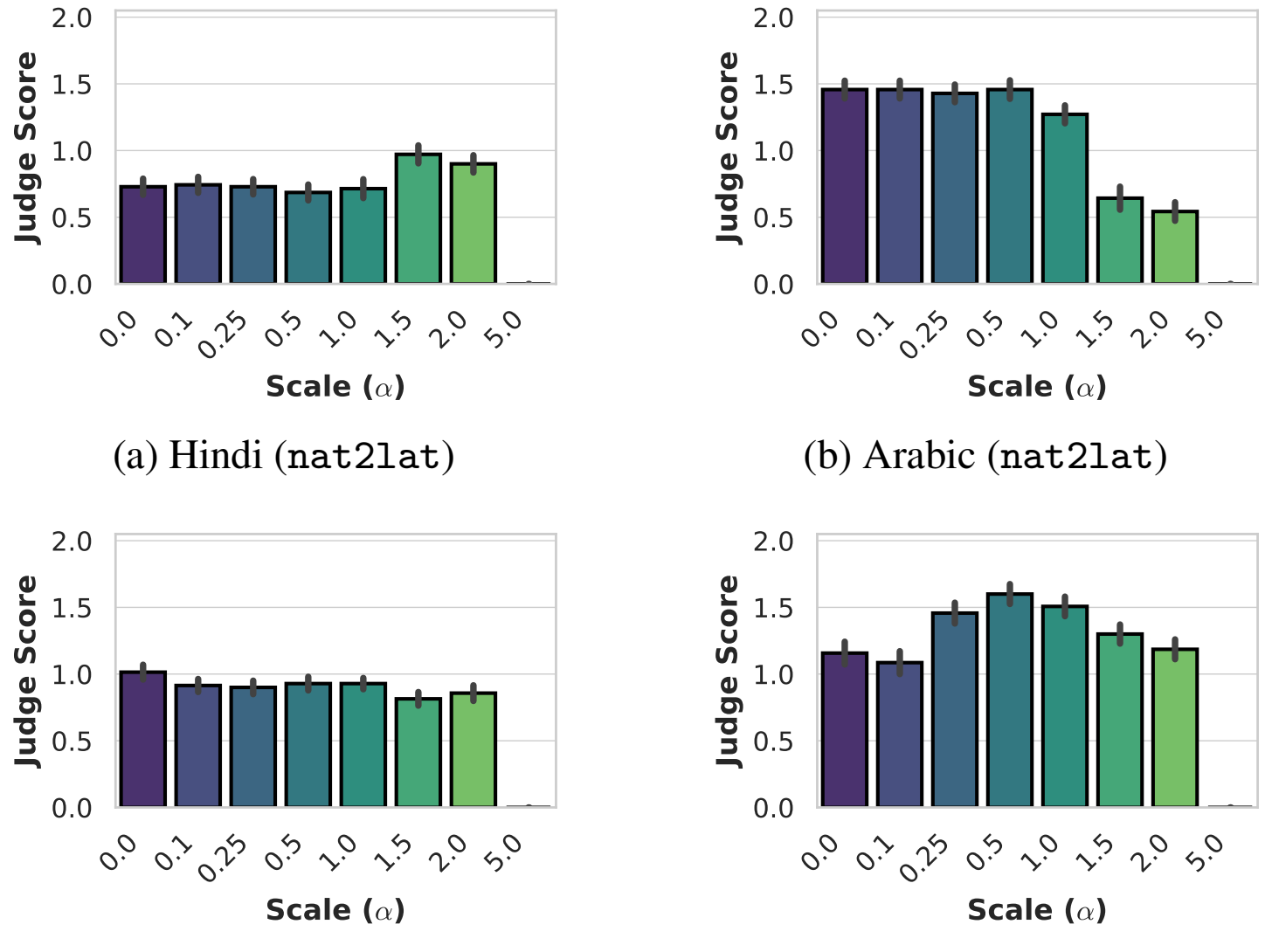

(a) Hindi (`nat2lat`) (b) Arabic (`nat2lat`)

(c) Hindi (`lat2nat`) (d) Arabic (`lat2nat`)

Figure 23: Response relevance after steering across languages for **`Aya-Expanse-32B`**. Results show the quality of steered responses across different alpha levels for Hindi and Arabic in both steering directions.

# C Mechanistic Analysis

## C.1 Activation Patching

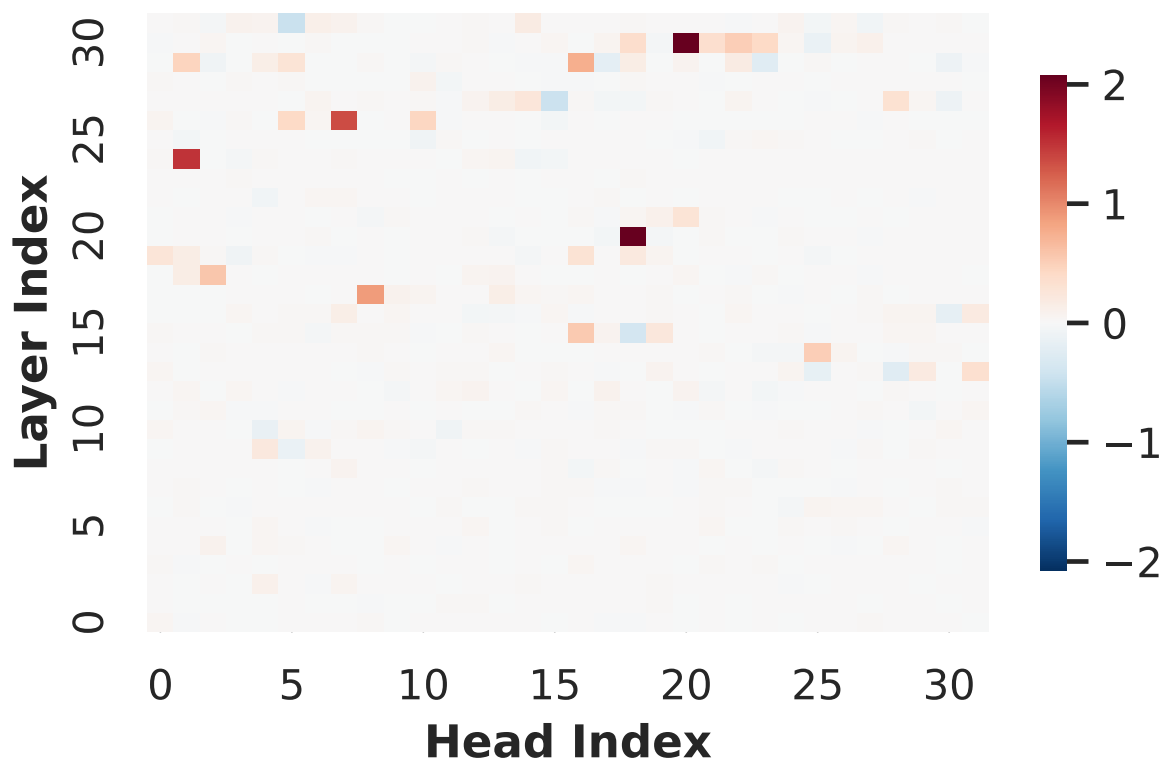

(a) Arabic: Native → Latin

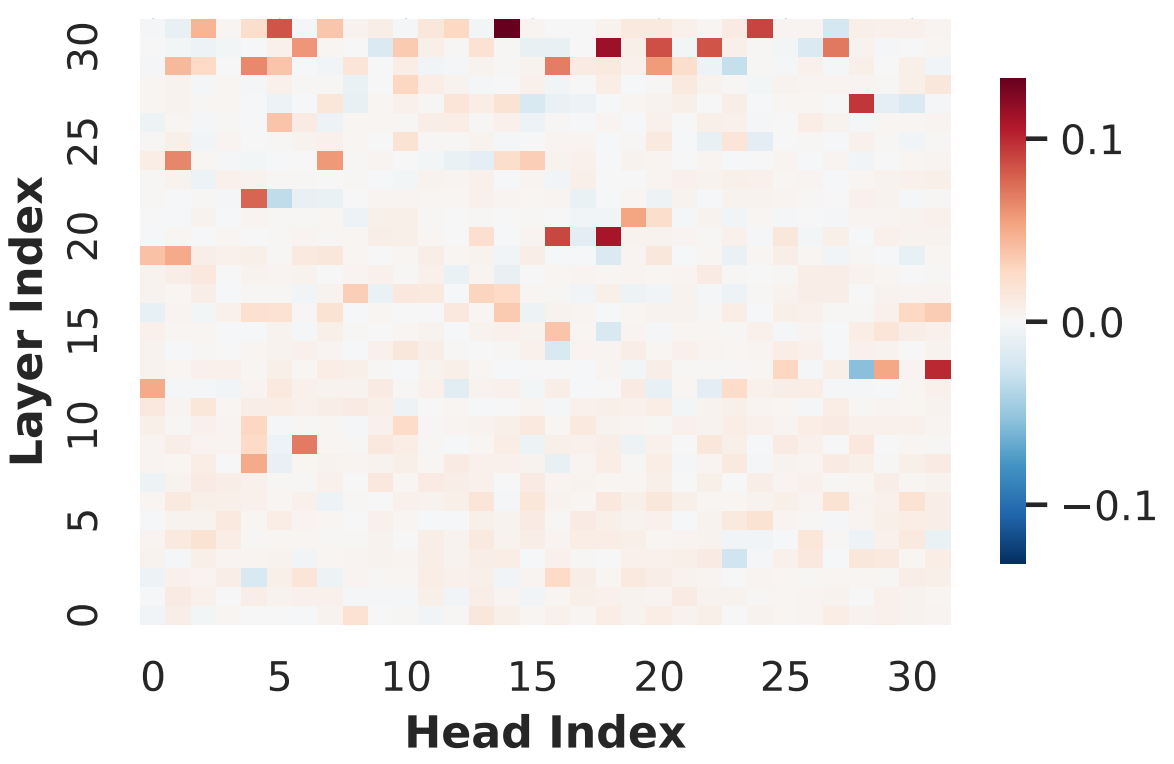

(b) Arabic: Latin → Native

Figure 24: Per-head causal contributions to script choice in `Llama-3.1-8B` for Arabic. Native-to-Latin shows concentrated effects in a small subset of heads, while Latin-to-native effects are more distributed and lower magnitude, indicating asymmetric control over script generation.

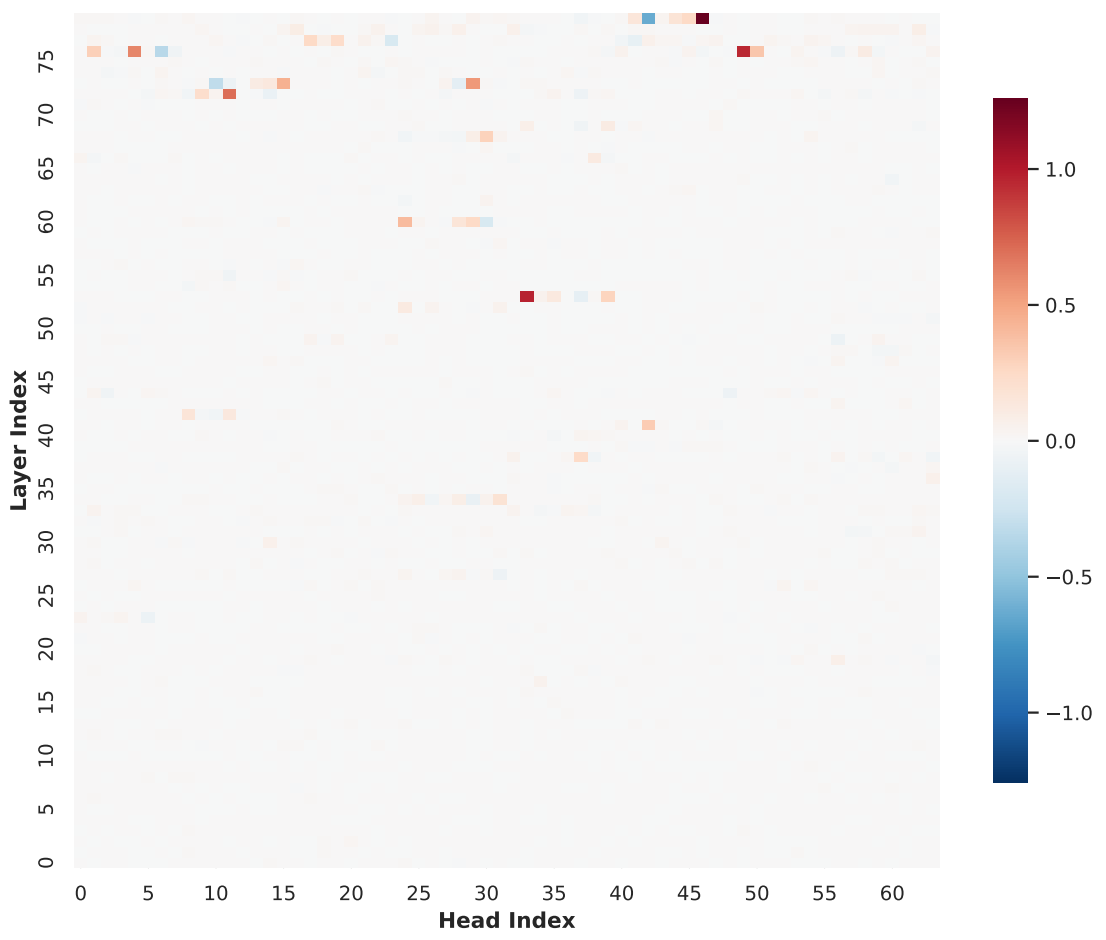

(a) Hindi: Native → Latin

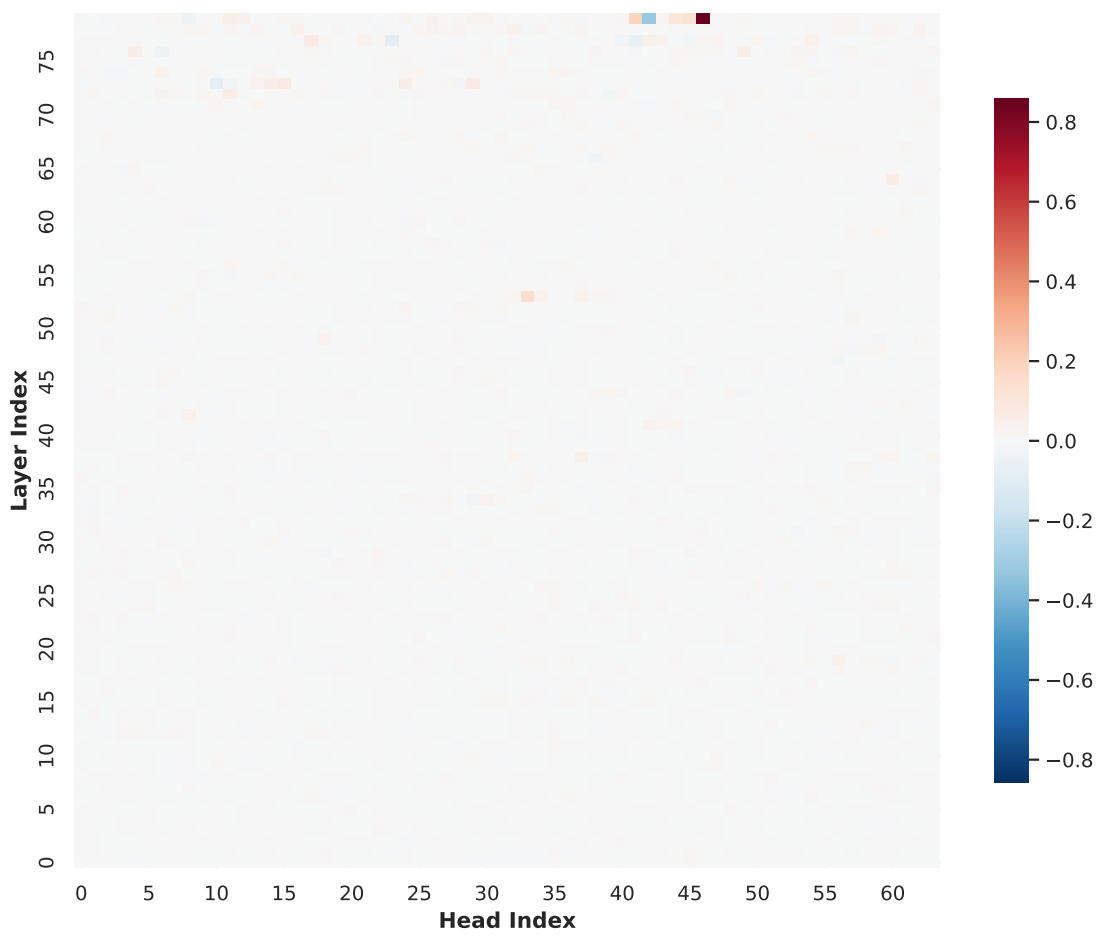

(b) Hindi: Latin → Native

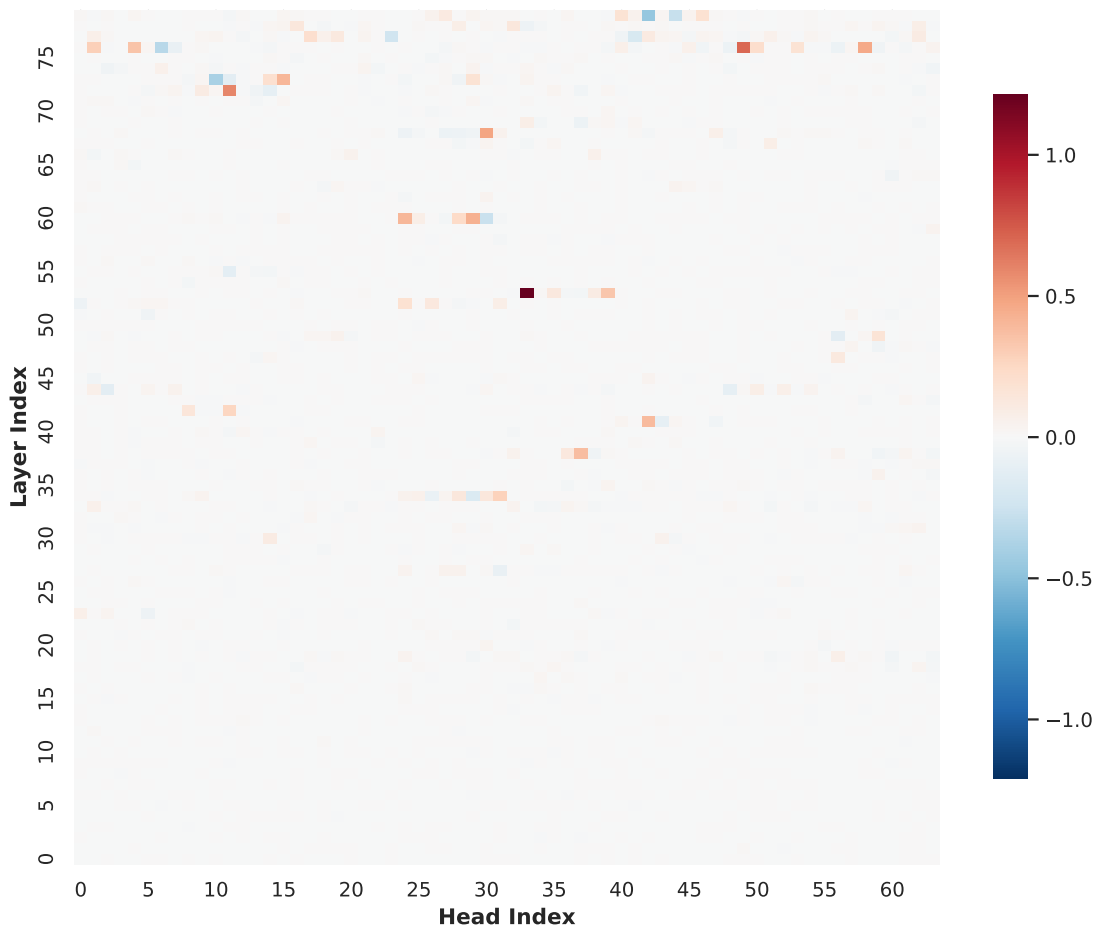

(c) Arabic: Native → Latin

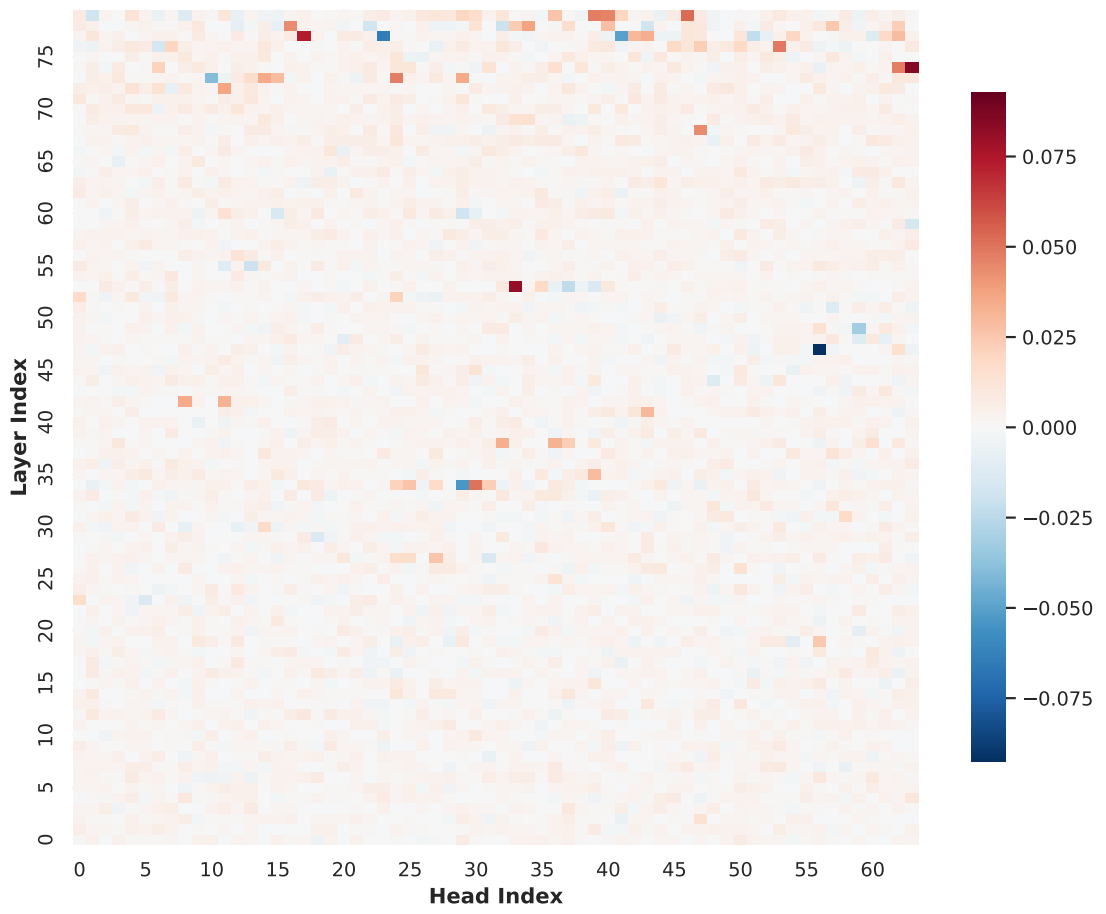

(d) Arabic: Latin → Native

Figure 25: Per-head causal contributions to script choice in `Llama-3.1-70B`.

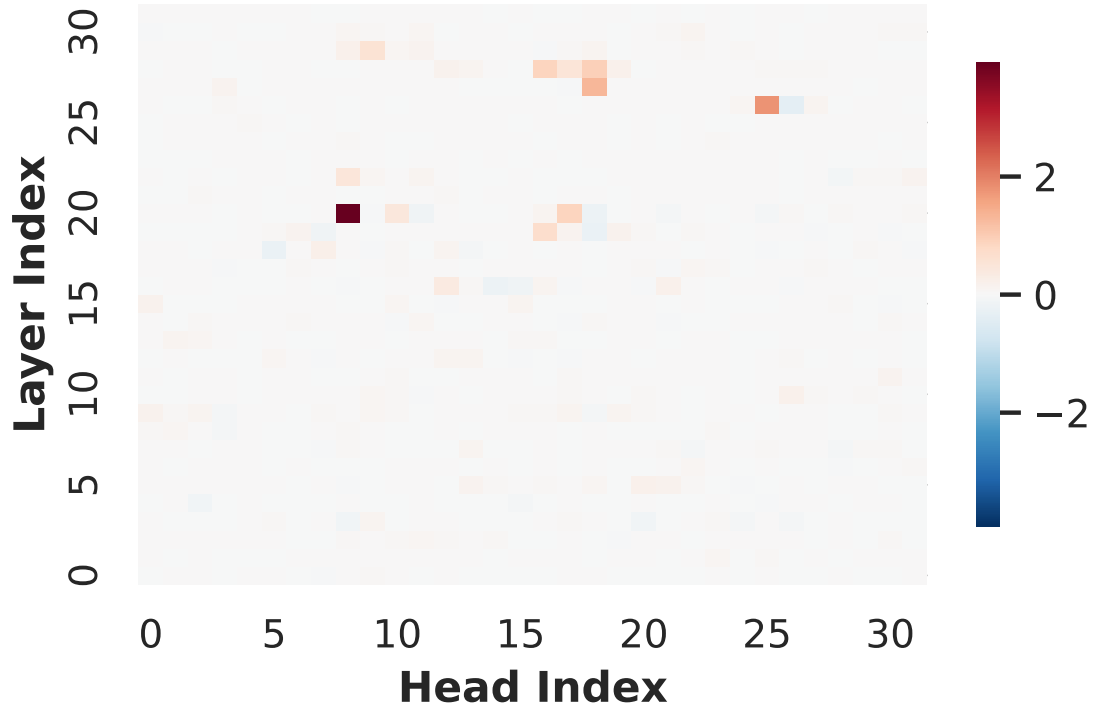

(a) Hindi: Native → Latin

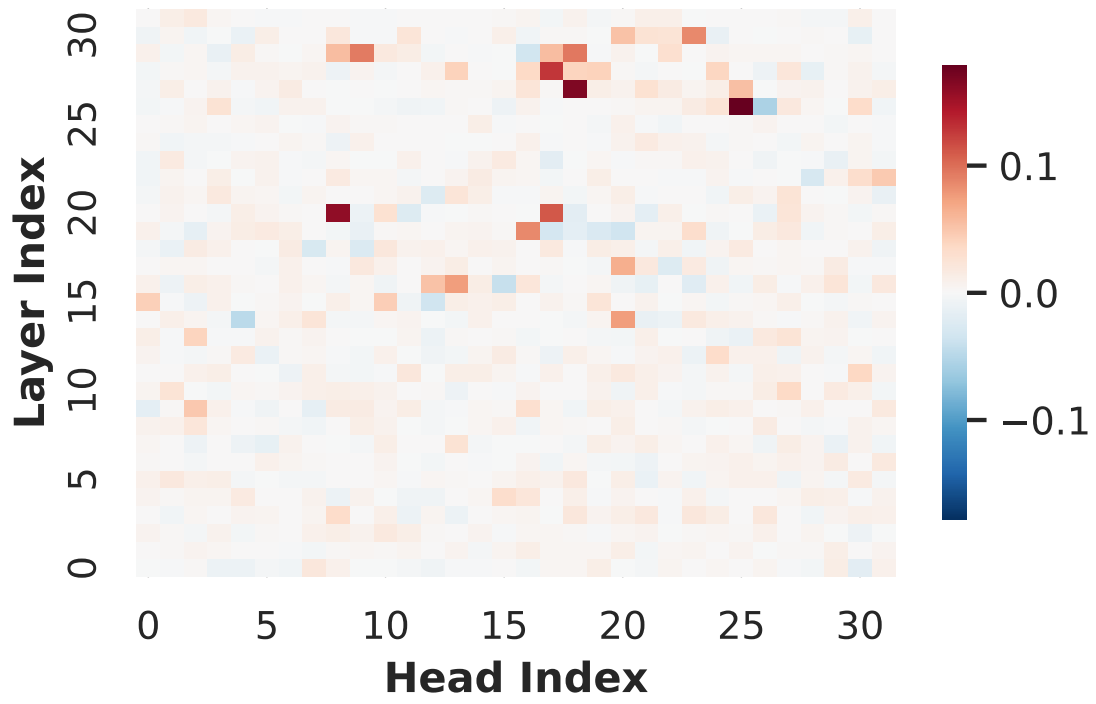

(b) Hindi: Latin → Native

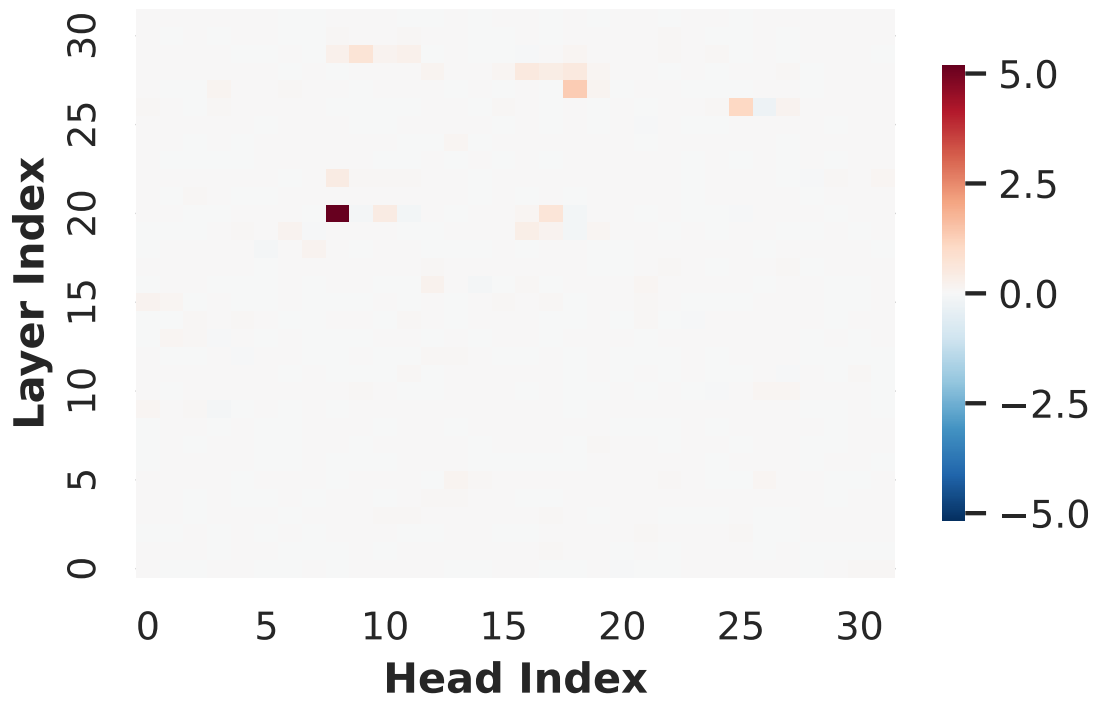

(c) Arabic: Native → Latin

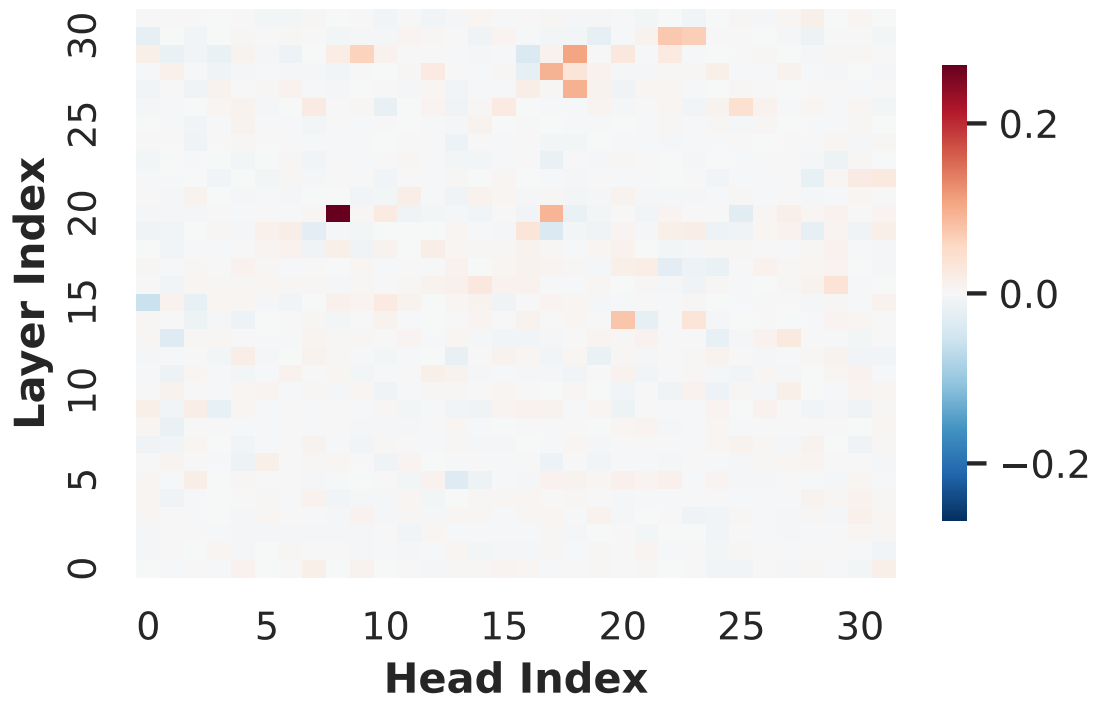

(d) Arabic: Latin → Native

Figure 26: Per-head causal contributions to script choice in `Aya-Expanse-8B`.

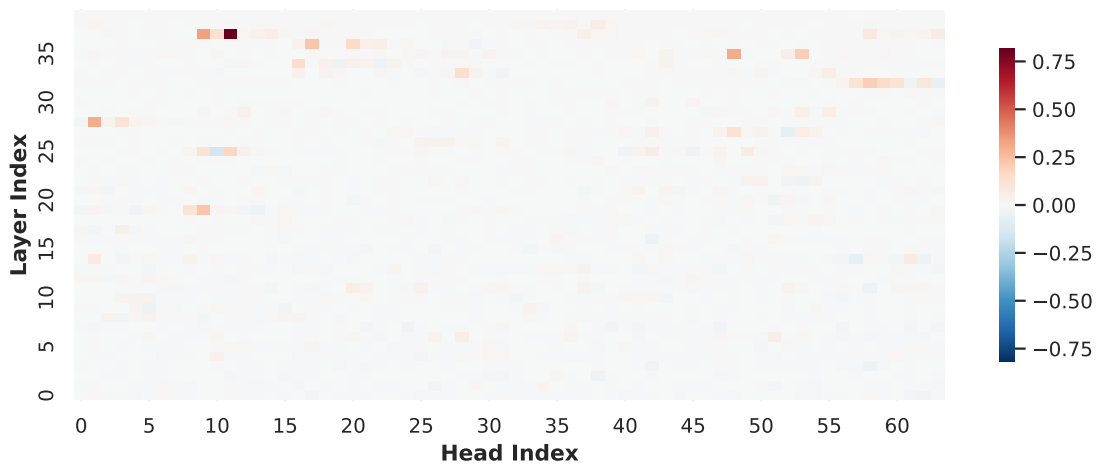

(a) Hindi: Native → Latin

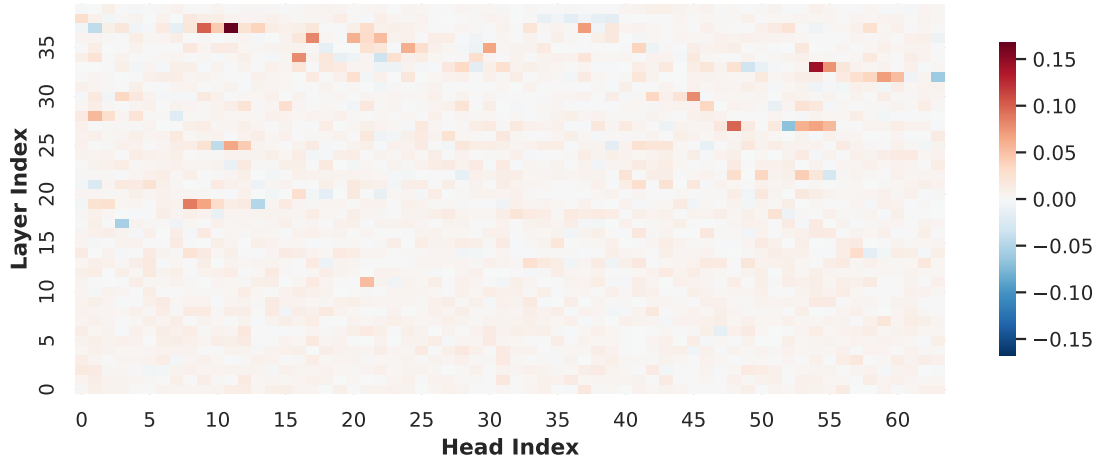

(b) Hindi: Latin → Native

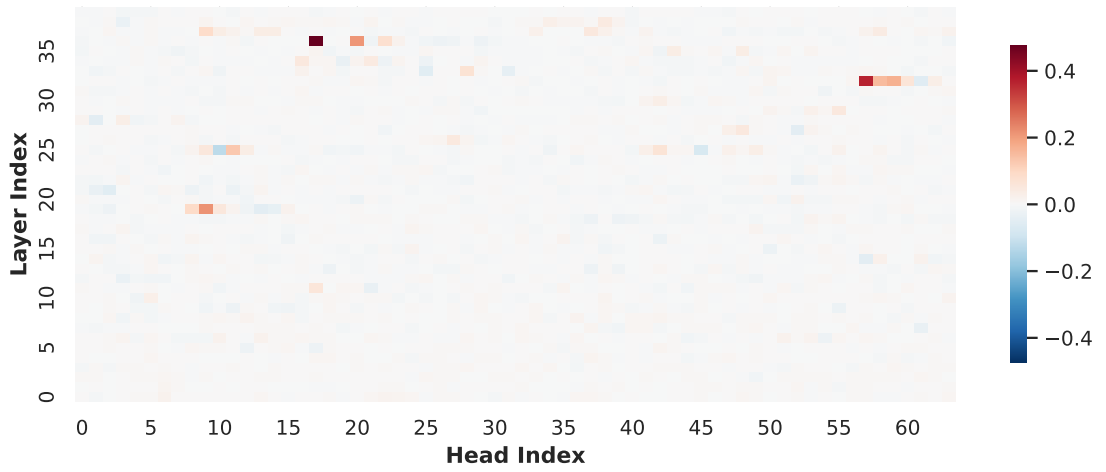

(c) Arabic: Native → Latin

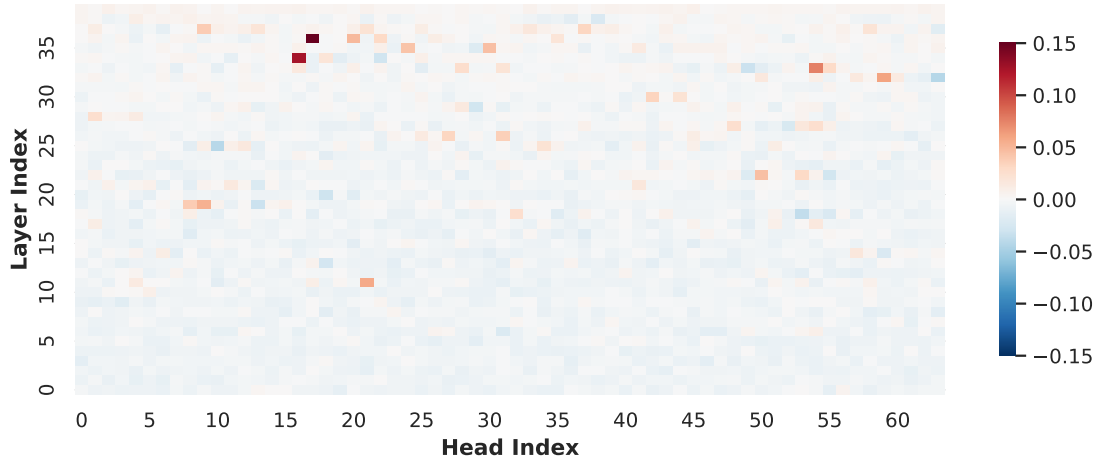

(d) Arabic: Latin → Native

Figure 27: Per-head causal contributions to script choice in `Aya-Expanse-32B`.

## C.2 Attention Head Validation

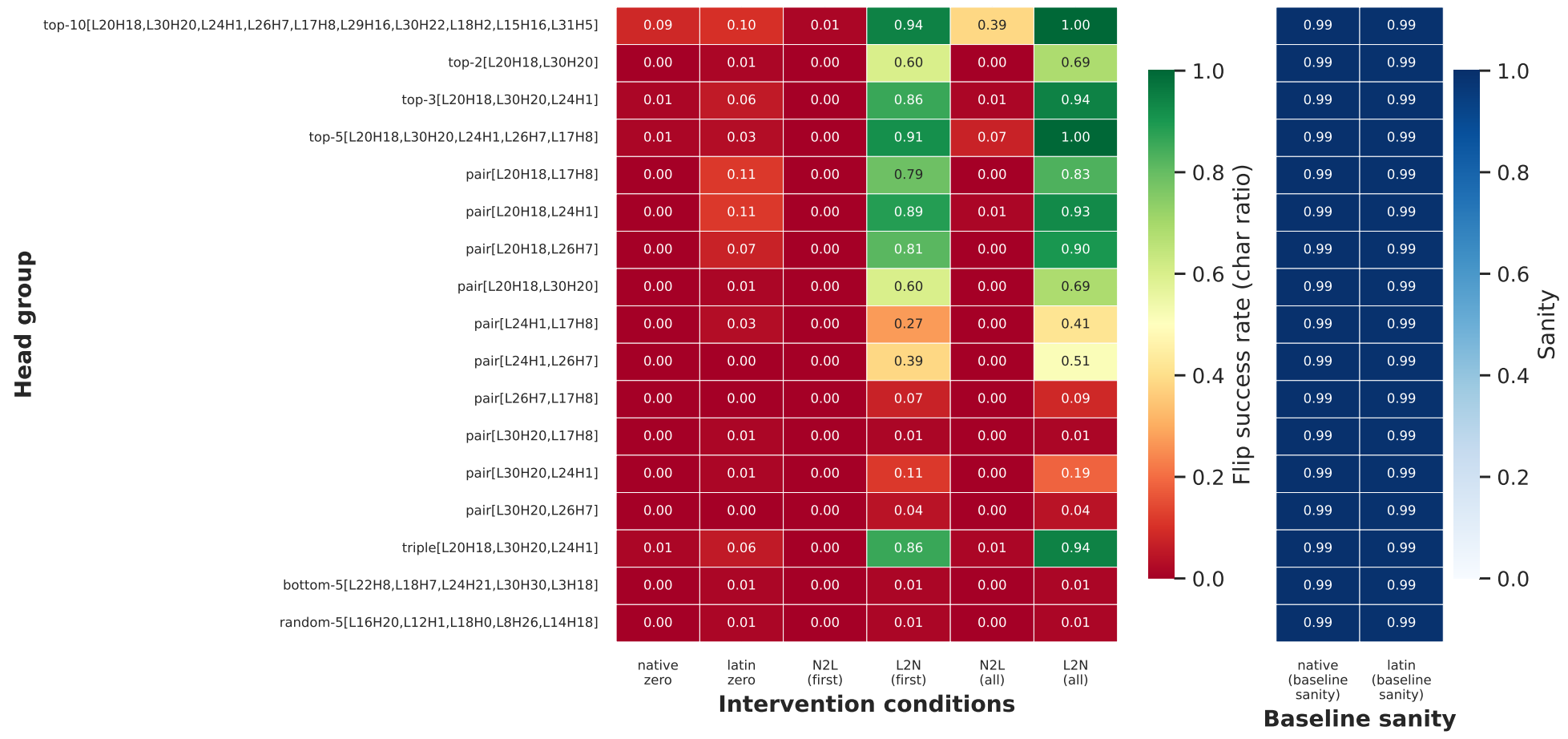

Figure 28: Validation of the identified script-mediating heads in `Llama-3.1-8B` on Arabic.

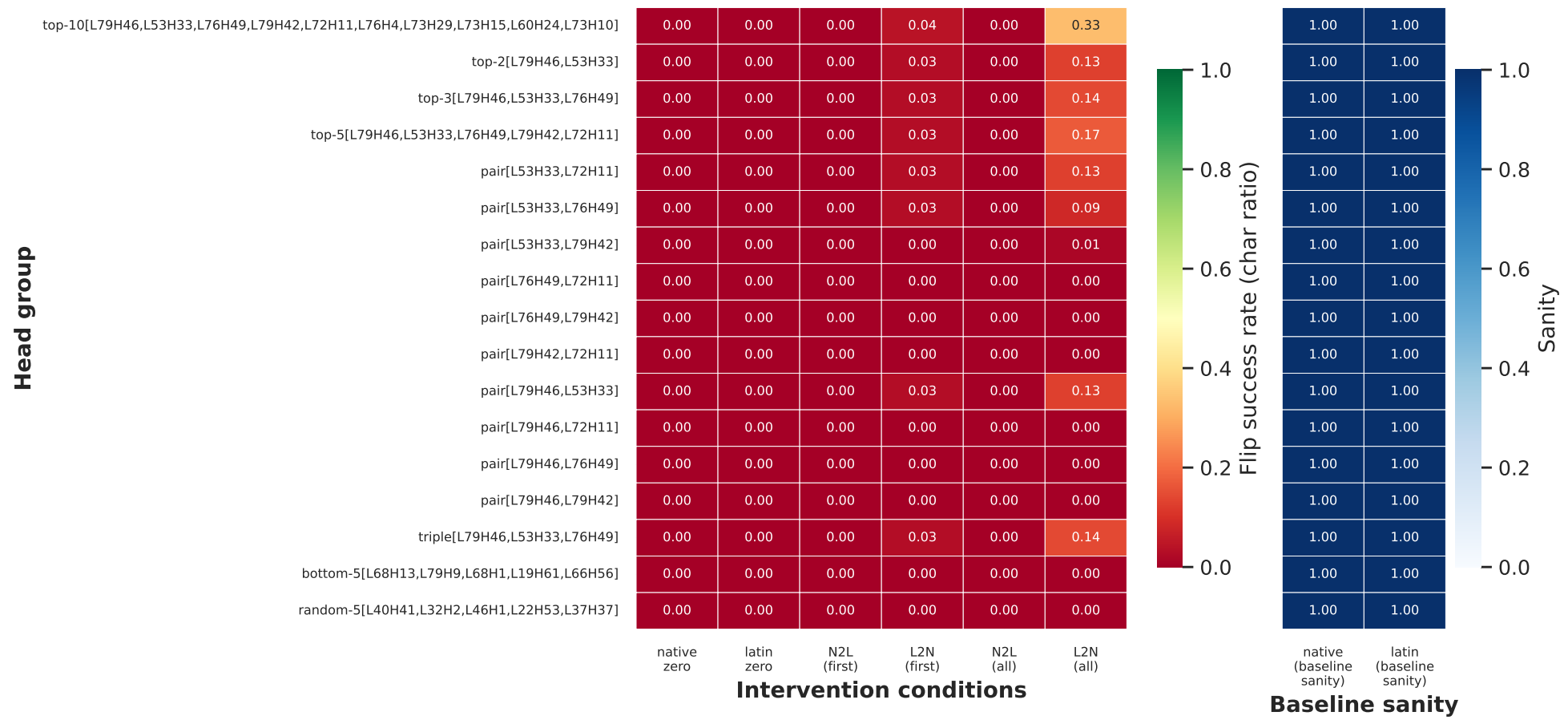

Figure 29: Validation of the identified script-mediating heads in `Llama-3.1-70B` on Hindi.[6]

[6]We note that larger models sometimes show weaker results, likely because the script mediation gate is distributed across more attention heads than the top-10 we restrict our analysis to.

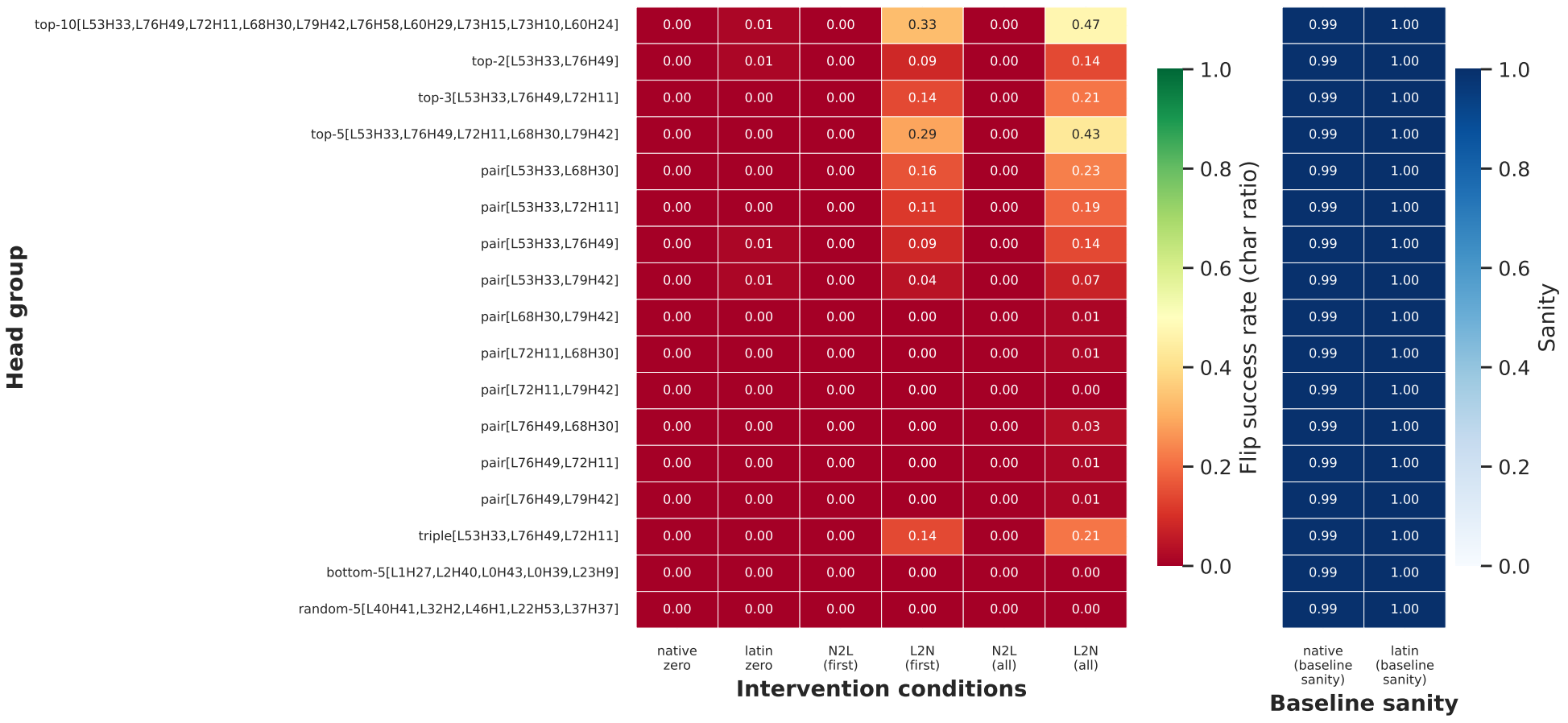

Figure 30: Validation of the identified script-mediating heads in `Llama-3.1-70B` on Arabic.

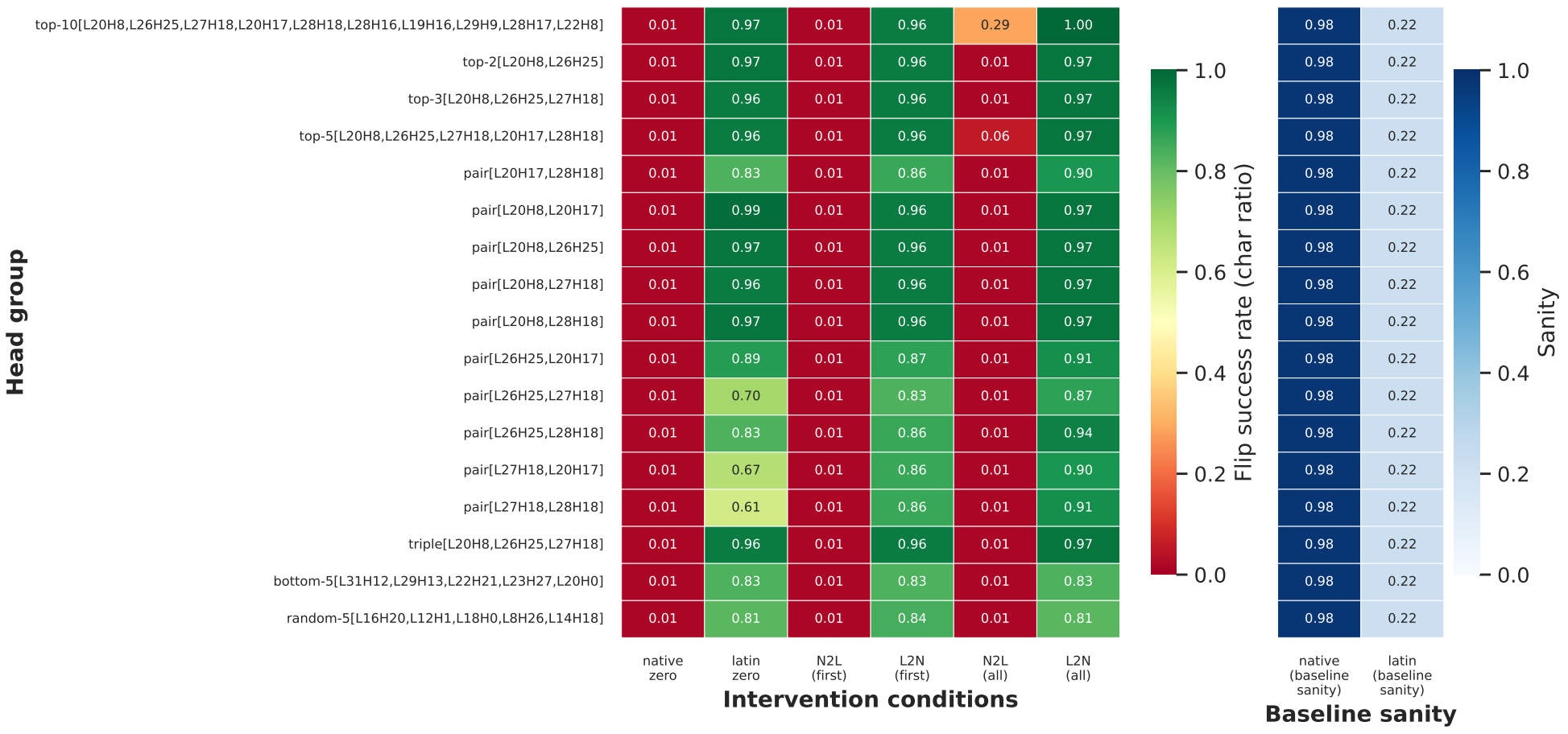

Figure 31: Validation of the identified script-mediating heads in `Aya-Expanse-8B` on Hindi.

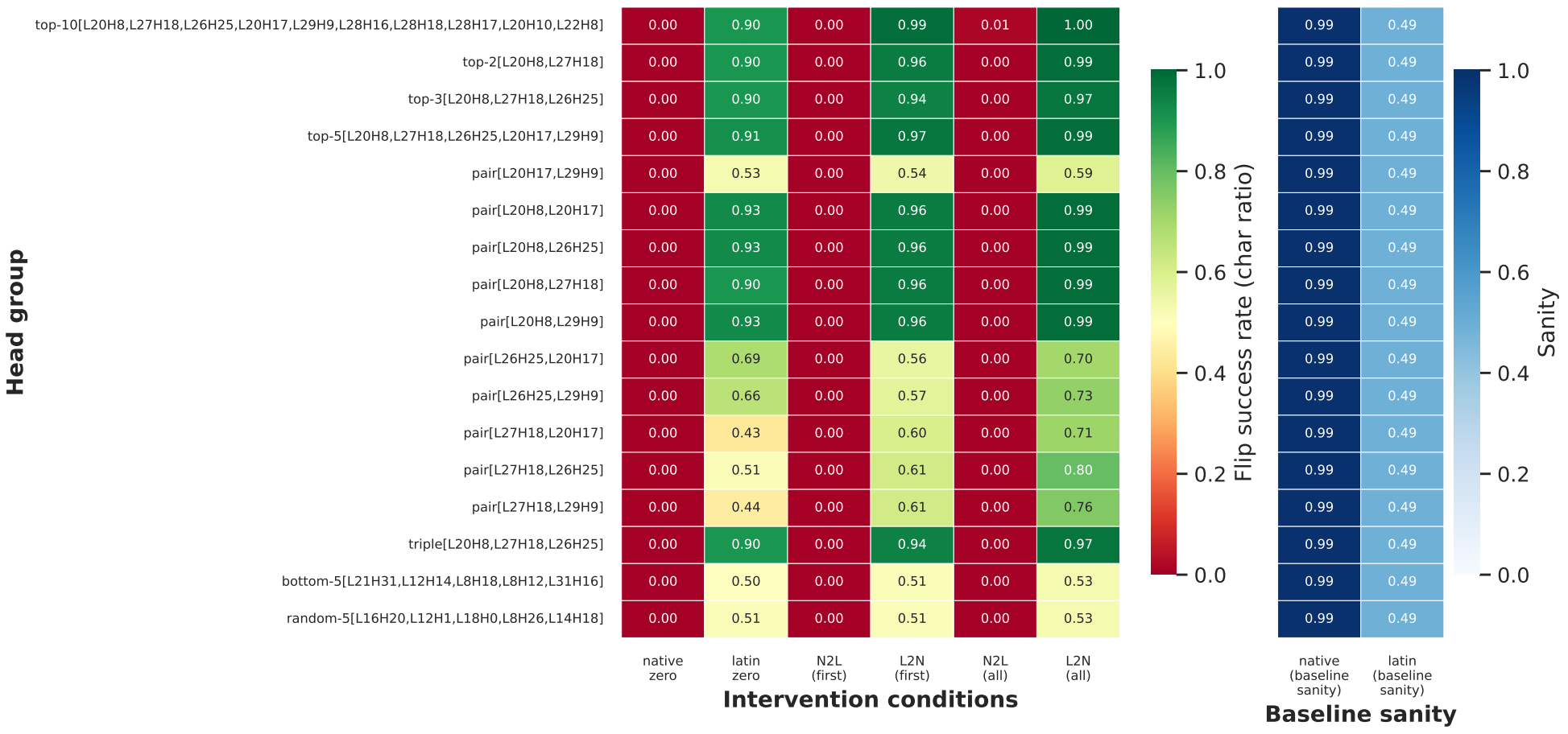

Figure 32: Validation of the identified script-mediating heads in `Aya-Expanse-8B` on Arabic.

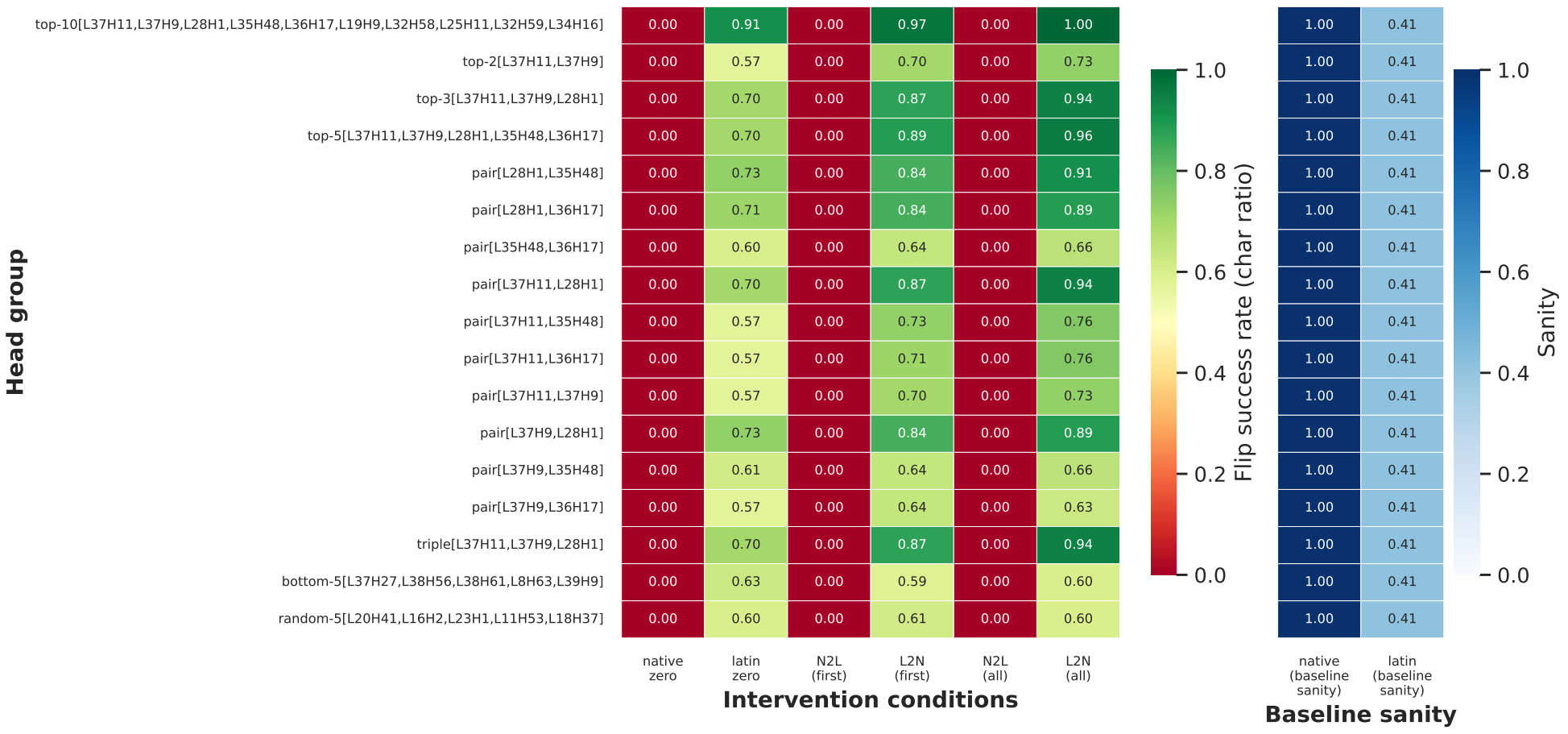

Figure 33: Validation of the identified script-mediating heads in `Aya-Expanse-32B` on Hindi.

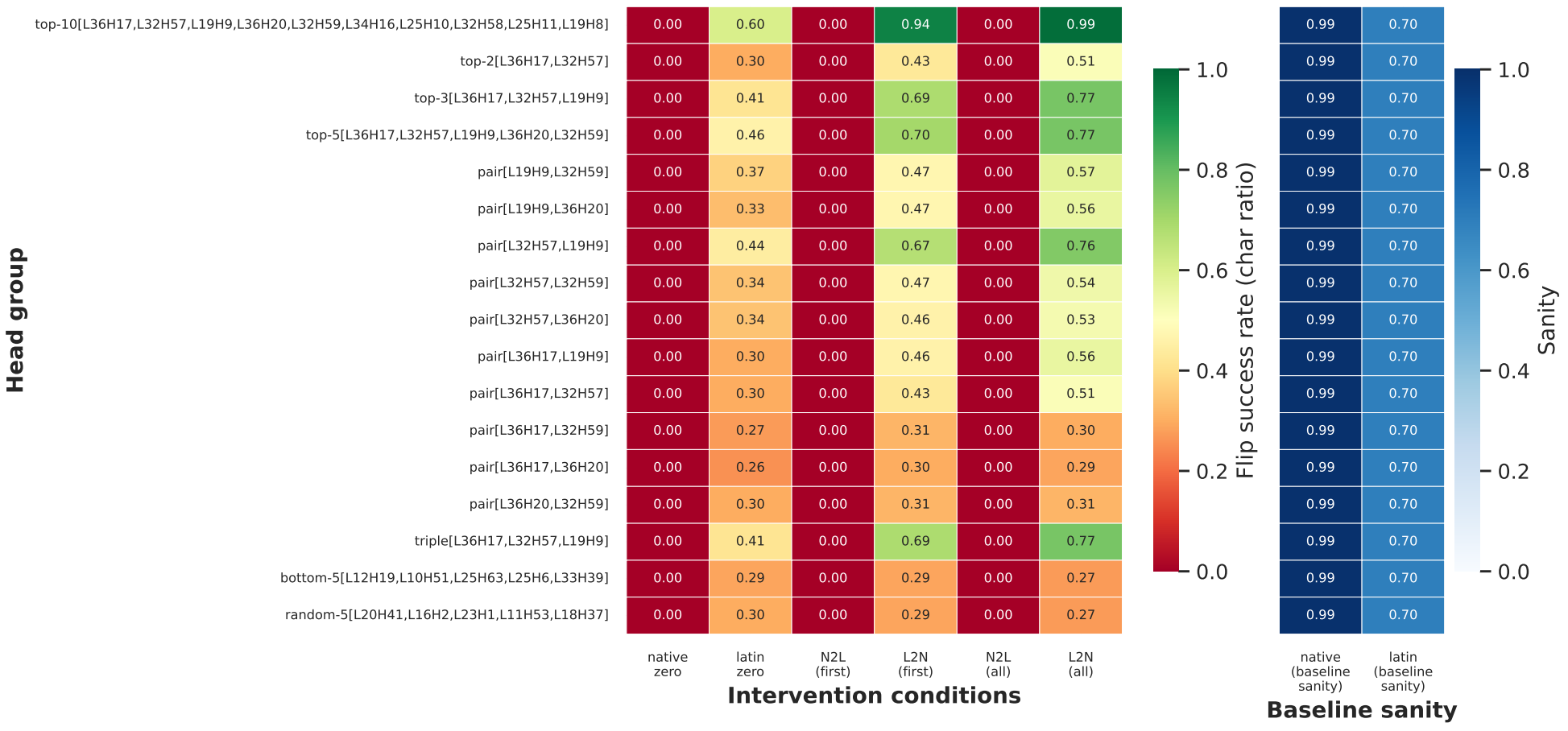

Figure 34: Validation of the identified script-mediating heads in `Aya-Expanse-32B` on Arabic.

### C.3 Attention Head Transfer

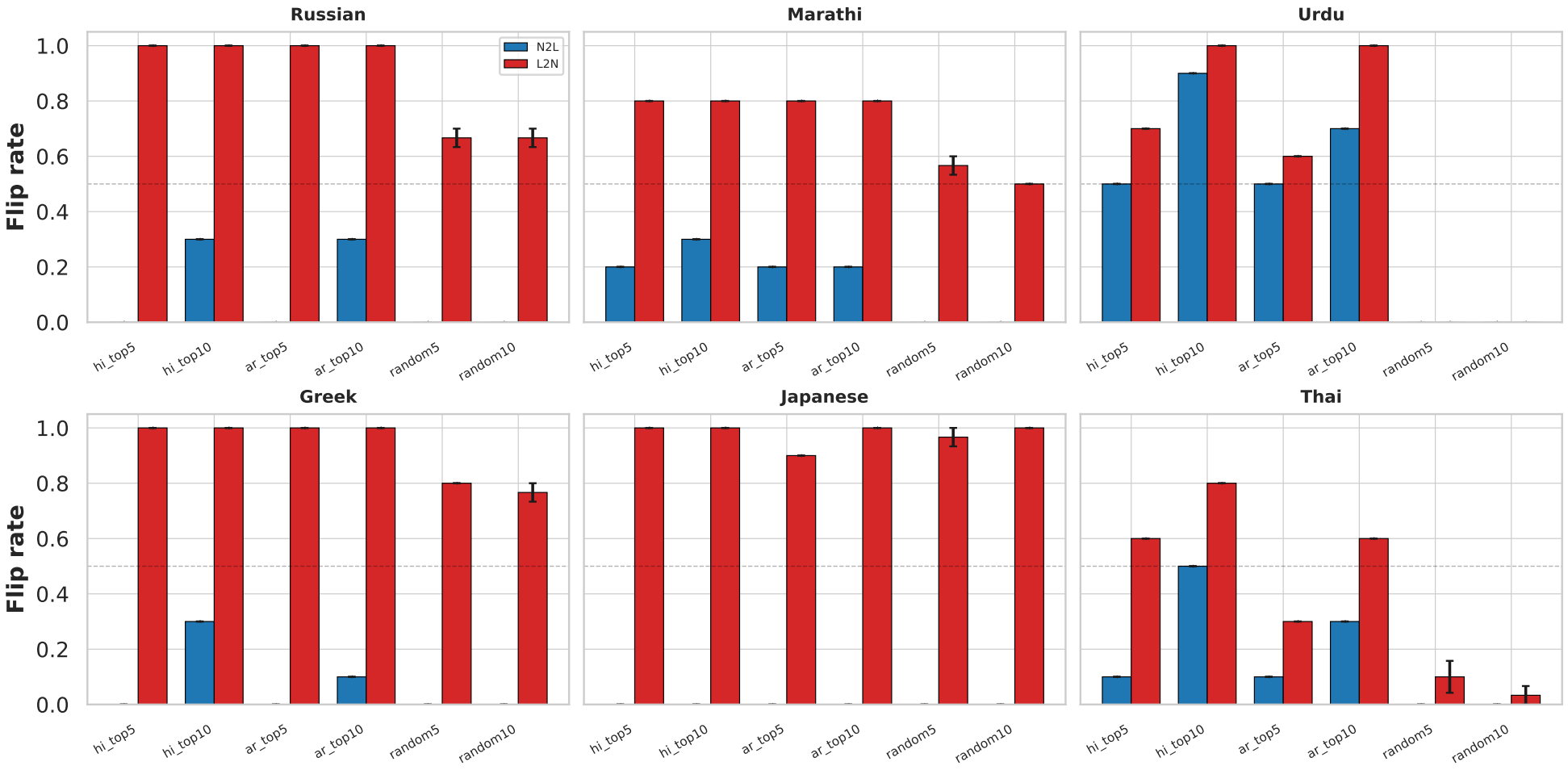

Figure 35: Applying identified script heads in `Aya-Expanse-8B` to other languages.

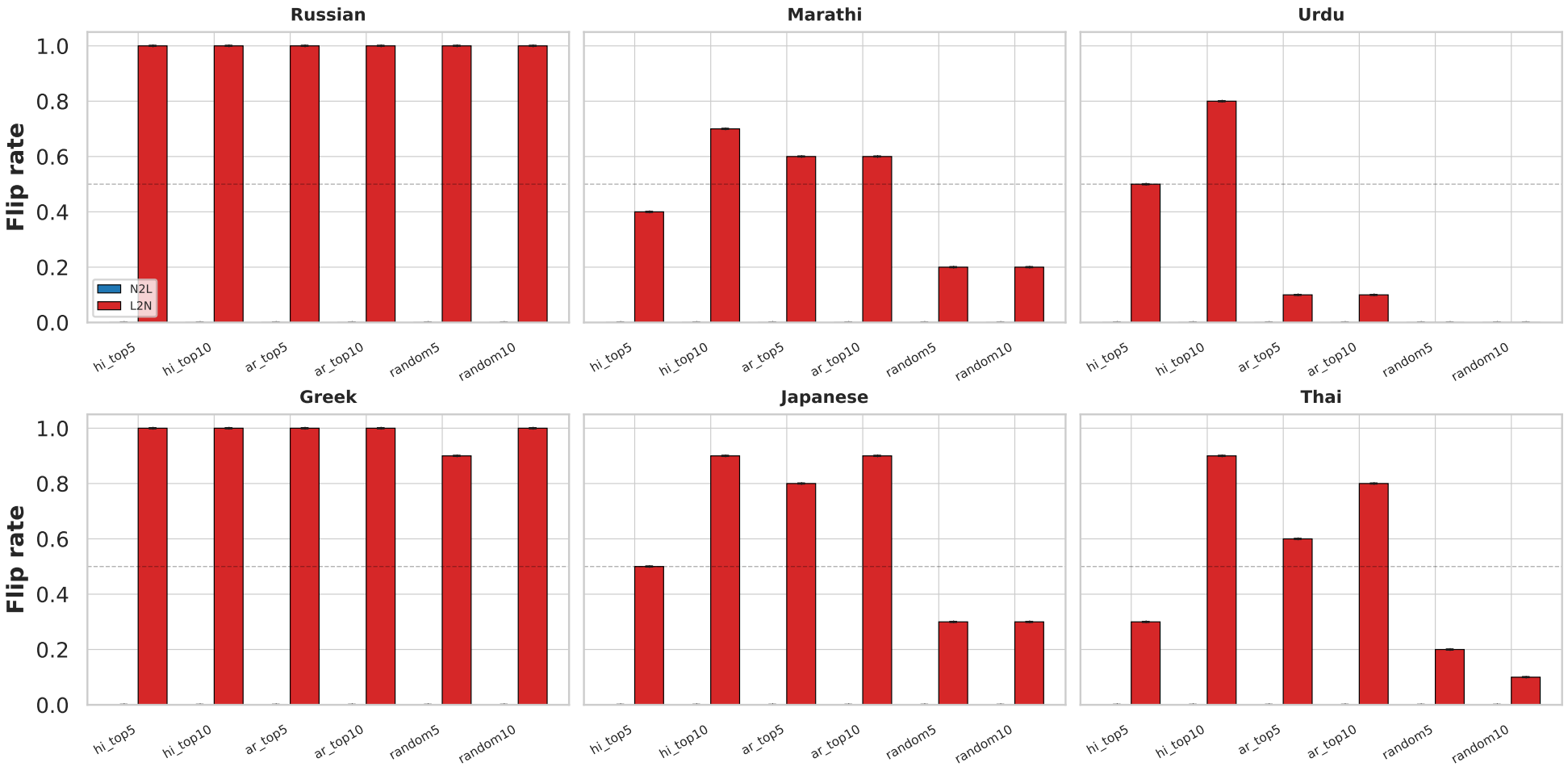

Figure 36: Applying identified script heads in `Aya-Expanse-32B` to other languages.

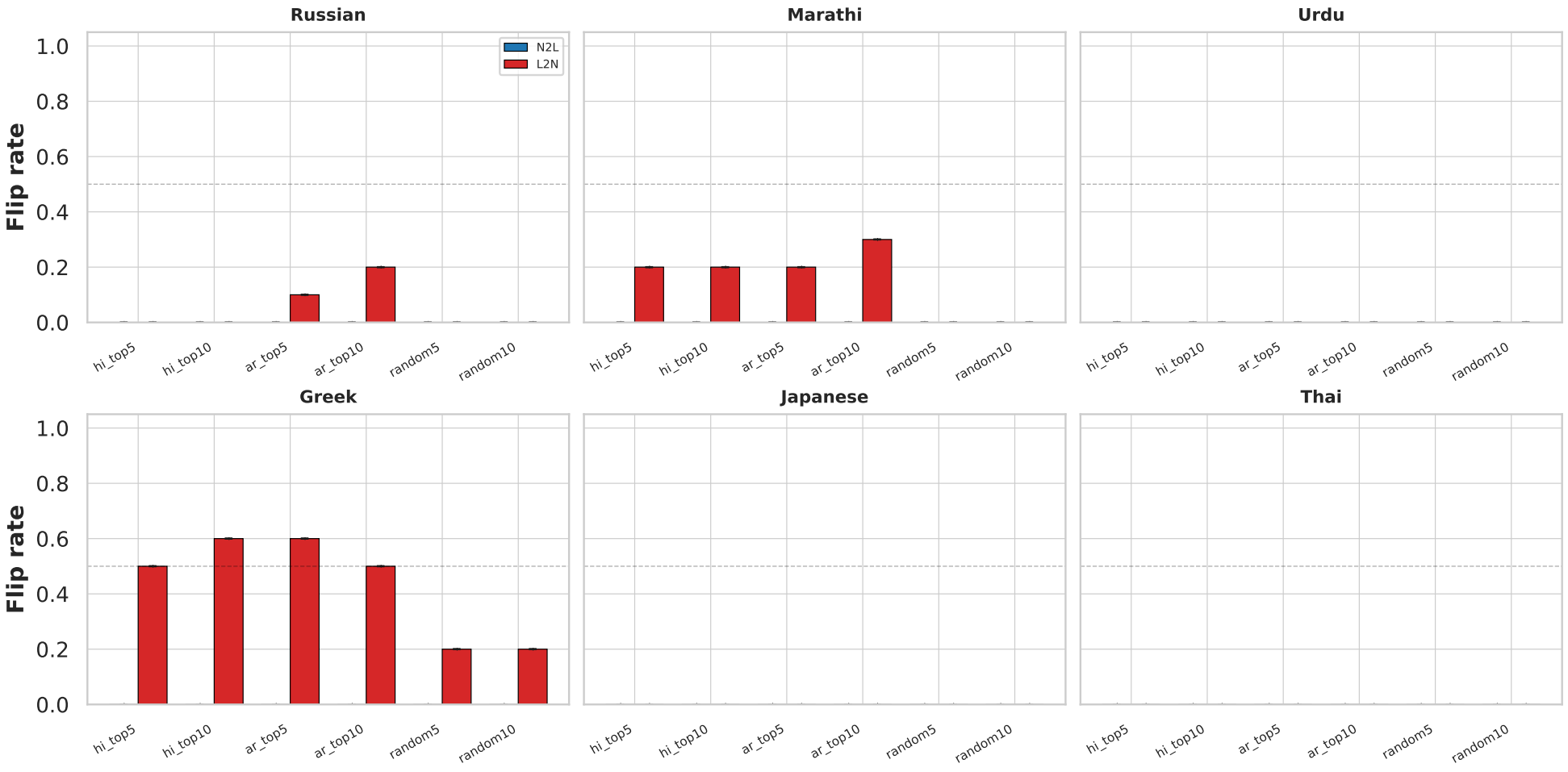

Figure 37: Applying identified script heads in `Llama-3.1-70B` to other languages.[7]

[7] Results are weaker here than for the smaller models, likely because the scrip gate is spread across more heads than the top-10 we restrict the analysis to.

### C.4 Qualitative Examples of Attention Head Transfer

| Lang. | Dir. | Prompt | Baseline | Patched (Top 10 heads) |
|---|---|---|---|---|
| **el** | N2L | Ποιο είναι το καλύτερο ελληνικό φαγητό; *(What is the best Greek food?)* | Είναι δύσκολο να καθορίσει κανείς το καλύτερο ελληνικό φαγητό, καθώς υπάρχουν πολλές εξαιρετικές επιλογές... | To "kalimavronto" einai mia apo tis megaliteres kleftikes kai traditikies sketes tis ellinikis gastronomias. To kalimavronto einai ena tipiko elliniko piato... |
| | L2N | Poio einai to kalytero elliniko fagito ? *(romanized)* | To "kalytero" einai enas oles tis ellinikis glossas synolikos oles tis lekseis pou einai kalyteres, alla to "kalytero elliniko fagito" einai enas tropos glossas... | Το καλύτερο ελληνικό φαγίτο είναι ένα θέμα που μπορεί να προκαλέσει συζητήσεις και διαφωνίες. Ωστόσο, κάποια από τα πιο δημοφιλή και εκλεκτά ελληνικά φαγίτα είναι... |
| **mr** | N2L | मला एक छोटी गोष्ट सांगा. *(Tell me a short story.)* | एक छोटी गोष्ट: एक छोटा पक्षी एक दिन एक विशाल झाडावर बसला होता. तो झाडावरून एक फळ खात होता... | Ek baar ek chhote sa gav mein ek chhota sa bachcha rehta tha. Uska naam Ravi tha. Ravi ek din apne pita ke saath... |
| | L2N | Mala ek chhoti gosht sanga. *(romanized)* | Aapke liye ek chhoti gosht sanga banane ka ek aisa vikalp hai jismein aapko sirf 5-6 ingredient ki aavashyakata hogi... | मीठी गोश्त संगा हे एक लोकप्रिय गुजराती व्याज आहे जे मुख्यत: गुजरातमध्ये आणि भारतातील इतर भागांमध्ये परिचित आहे... |
| **th** | N2L | อธิบายเกี่ยวกับภูเขาดอยอินทนนท์ | ภูเขาดอยอินทนนท์<br>เป็นภูเขาในประเทศไทย<br>ตั้งอยู่ในจังหวัดเชียงใหม่<br>เป็นภูเขาที่สูงที่สุดในประเทศไทย... | ภูเขาดอยอินทนนท์<br>เป็นภูเขาในประเทศไทย<br>ตั้งอยู่ในจังหวัดเชียงใหม่<br>เป็นหนึ่งในภูเขาที่มีความสำคัญ...<br>[No script flip] |
| | L2N | Athibai kiao kap phukhao Doi Inthanon. *(romanized)* | Athibai kiao kap phukhao Doi Inthanon ni khao phra that hai nai changwat Chiang Mai, Thailand... | คุณกำลังถามถึงสถานที่ที่มีชื่อเสียงของประเทศไทย คือ "ดอยอินทนนท์" เป็นภูเขาและจุดชมวิว... |

Table 2: Examples of patching the top-10 heads identified on Arabic, which causally flips the output script in `Llama-3.1-8B`. Baseline outputs preserve the input script, while patched outputs switch to the opposite script while maintaining coherent content. We omit results for other languages due to challenges in integrating multiple scripts within LaTeX typesetting; full results are available in our GitHub repository.